%% file: main.tex
\documentclass[conference]{IEEEtran}
\usepackage{times}

\usepackage[numbers]{natbib}
\usepackage{multicol}
\usepackage[bookmarks=true]{hyperref}

\input{includes}

\begin{document}

\title{Dynamic Rank Adjustment in Diffusion Policies for Efficient and Flexible Training}


\author{\authorblockN{Xiatao Sun$^{1\dagger}$,
Shuo Yang$^2$,
Yinxing Chen$^1$, 
Francis Fan$^1$,
Yiyan (Edgar) Liang$^2$,
Daniel Rakita$^{1\dagger}$,
\authorblockA{$^1$Yale University \quad $^2$University of Pennsylvania}
\authorblockA{$^\dagger$Corresponding author \quad \{xiatao.sun, daniel.rakita\}@yale.edu}
}}

\maketitle

\input{0-abstract}
\input{1-introduction}

\input{2-background}

\input{3-related_works}
\input{4-drift_framework}

\input{5-drift_dagger}

\input{6-simulation_evaluation}
\input{7-real_world_evaluation}
\input{8-discussion}

\IEEEpeerreviewmaketitle

\section*{Acknowledgments}
This work was supported by Office of Naval Research award N00014-24-1-2124


\bibliographystyle{plainnat}
\bibliography{references}

\input{9-appendix}

\end{document}

%% file: includes.tex
\usepackage{times}
\usepackage[numbers]{natbib}
\usepackage{multicol}
\usepackage[bookmarks=true]{hyperref}
\usepackage{lipsum}
\usepackage{comment}
\usepackage{balance}
\usepackage{graphicx}
\usepackage[table]{xcolor}
\usepackage{listings}
\usepackage{amsfonts}
\usepackage{amsmath}
\usepackage{amsthm}
\usepackage[linesnumbered,ruled,noend,vlined]{algorithm2e}

\SetCommentSty{mycommfont}

\usepackage{seqsplit}

\theoremstyle{definition}

\usepackage{array}
\usepackage{tabularx} 
\usepackage{booktabs}
\usepackage{makecell}
\usepackage{multirow}
\usepackage[labelformat=empty]{subcaption}

\usepackage{caption}
\usepackage{amssymb}

%% file: 0-abstract.tex
\begin{abstract}

Diffusion policies trained via offline behavioral cloning have recently gained traction in robotic motion generation. While effective, these policies typically require a large number of trainable parameters. This model size affords powerful representations but also incurs high computational cost during training. Ideally, it would be beneficial to \textit{dynamically adjust} the trainable portion as needed, balancing representational power with computational efficiency. For example, while overparameterization enables diffusion policies to capture complex robotic behaviors via offline behavioral cloning, the increased computational demand makes online interactive imitation learning impractical due to longer training time. To address this challenge, we present a framework, called DRIFT, that uses the Singular Value Decomposition to enable dynamic rank adjustment during diffusion policy training. We implement and demonstrate the benefits of this framework in DRIFT-DAgger, an imitation learning algorithm that can seamlessly slide between an offline bootstrapping phase and an online interactive phase. We perform extensive experiments to better understand the proposed framework, and demonstrate that DRIFT-DAgger achieves improved sample efficiency and faster training with minimal impact on model performance. The project website is available at: \href{https://apollo-lab-yale.github.io/25-RSS-DRIFT-website/}{https://apollo-lab-yale.github.io/25-RSS-DRIFT-website/}.

\end{abstract}


%% file: 1-introduction.tex
\section{Introduction}
\label{sec:introduction}

Diffusion policies have recently emerged as a powerful paradigm for robotic motion generation, demonstrating impressive performance across various manipulation tasks \cite{chi_dp}. Their strong performance is attributed to overparameterization, which has been shown to enhance both the generalization and representational capacity of neural networks \cite{dar_overparam}. However, this advantage comes with a significant drawback, which is the inefficiency of training overparameterized diffusion models \cite{zhang_overparam_dm}. While the machine learning community has made strides in addressing this inefficiency by leveraging the intrinsic low-rank structure of neural networks \cite{li_low_rank, li_low_rank2}, these approaches primarily focus on fine-tuning pre-trained models \cite{hu_lora, dettmers_qlora}. In robotics, however, policies are often trained from scratch, making these fine-tuning methods unsuitable.

To exploit the intrinsic low-rank structure, training from scratch in robotics necessitates a dynamic approach to balancing the trade-off between overparameterization, which provides strong representational power, and the efficiency gained from reduced-rank training. Ideally, at the beginning of training, maintaining high trainable ranks allows the policy to capture the general patterns of desired behaviors. As training progresses, the number of trainable ranks can be reduced to improve training efficiency, as the policy shifts to incremental refinement. This capability is particularly valuable in scenarios like interactive imitation learning (IL) with diffusion policies.

\begin{figure}
\centering
\includegraphics[width=\columnwidth]{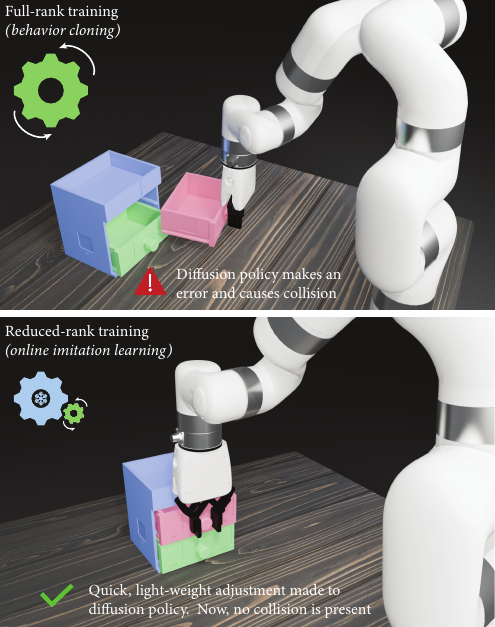}
\caption{This paper explores balancing overparameterization and training efficiency in diffusion policies by dynamically adjusting the frozen and trainable portions of weight matrices. In the top section of the figure, the learner, trained offline via behavior cloning with full-rank training, attempts to insert the upper drawer box into the container but fails due to collisions with both the container and the lower drawer box. In the bottom section, after efficient online adaptation with reduced trainable ranks, the learner efficiently improves its performance, successfully completing the task.}
\label{fig:teaser_img}
\end{figure}

Before the adoption of diffusion policies, interactive IL methods typically used simple and compact network architectures as policy backbones. These methods were developed to address the sample inefficiency of behavior cloning (BC) \cite{sun_il, zare_il_survey}, and often involve an offline bootstrapping stage for initial training, followed by an online adaptation stage where experts provide corrective interventions to refine the policy. However, directly extending these methods to diffusion policies is impractical due to their significantly larger number of trainable parameters, which is often an order of magnitude greater than those of compact networks and results in substantially increased training times. This challenge undermines the feasibility of online interactive IL with diffusion policies.

Typically, diffusion policies are trained offline via BC, where a large dataset of demonstrations is collected, and training occurs in isolation. However, when these policies underperform, the expert must collect additional demonstrations targeting the challenging trajectories, provide corrective demonstrations, and retrain the policy iteratively in an offline manner. This process is both inefficient and unintuitive, as the expert often has limited insight into the trajectories where the policy struggles and may find it difficult to reproduce such challenging scenarios.

To address these limitations, we propose \textbf{D}ynamic \textbf{R}ank-adjustable D\textbf{IF}fusion Policy \textbf{T}raining (DRIFT), a framework designed to enable dynamic adjustments of the number of trainable parameters in diffusion policies through reduced-rank training. The framework introduces two key components, which are \emph{rank modulation} that leverages Singular Value Decomposition (SVD) \cite{strang2022introduction} to adjust the proportion of trainable and frozen ranks while maintaining the total rank constant, and \emph{rank scheduler} that dynamically modulates the number of trainable ranks during training using a decay function.


\begin{figure}
\centering
\includegraphics[width=\columnwidth]{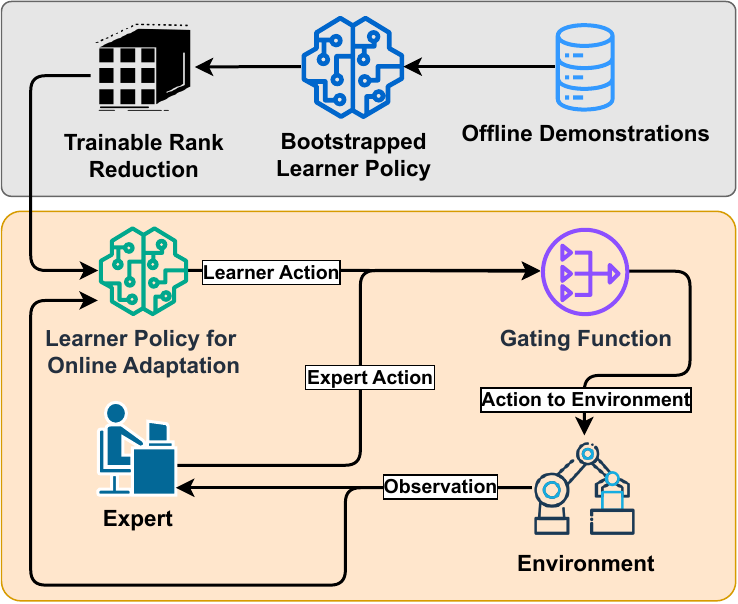}
\caption{DRIFT-DAgger combines offline policy bootstrapping with online adaptation. The gating function, following the nomenclature of HG-DAgger \cite{kelly_hg_dagger}, refers to expert intervention and demonstration when the learner reaches undesirable states during online adaptation. Compared to BC, DRIFT-DAgger reduces the need for expert labeling while maintaining high performance. The trainable rank reduction accelerates batch training, improving the usability and practicality of online adaptation for large models without sacrificing performance.}
\label{fig:workflow}
\end{figure}

To demonstrate and evaluate the effectiveness of DRIFT, we instantiate and implement it into DRIFT-DAgger, an expert-gated interactive IL method that incorporates reduced-rank training. As shown in Fig. \ref{fig:workflow}, DRIFT-DAgger uses low-rank component to speed up training of diffusion policies. By freezing a significant portion of the ranks during online adaptation, DRIFT-DAgger reduces training time, making online interactive IL with diffusion policies more practical.

Despite being inspired by existing parameter-efficient fine-tuning methods, DRIFT-DAgger with rank modulation and rank scheduler is specifically designed for training diffusion policies from scratch and dynamically adjusting the trainable ranks, avoiding the need to reinitialize and inject low-rank components during training. This design enhances stability and reduces the time for forward passes during each training batch (\S\ref{sec:abl_rm}). Additionally, we perform extensive ablation studies on different variants of rank schedulers (\S\ref{sec:abl_rs}), and minimum trainable ranks (\S\ref{sec:abl_tr}). By combining diffusion policies with online interactive IL, DRIFT-DAgger improves sample efficiency compared to training diffusion policies with BC (\S\ref{sec:sim_exp}). We also validate our methods in real-world scenarios (\S\ref{sec:real_exp}). Finally, we discuss the limitations and implications of our work (\S\ref{sec:discussion}). Our contributions are as follows:
\begin{itemize}
    \item We propose DRIFT, a framework for diffusion policies that includes rank modulation and rank scheduler as novel components that exploit the intrinsic low-rank structure of overparameterized models, balancing training efficiency and model performance.
    \item We instantiate DRIFT into DRIFT-DAgger, an interactive IL algorithm that combines offline bootstrapping with efficient online adaptation, enabling effective integration of expert feedback during the novice policy training.
    \item We perform extensive experiments to demonstrate that DRIFT-DAgger improves sample efficiency and reduces training time while achieving comparable performance to diffusion policies trained with full rank.
    \item We provide open-source implementations in Pytorch for the DRIFT framework and DRIFT-DAgger algorithm.\footnote{\href{https://github.com/Apollo-Lab-Yale/drift_dagger}{https://github.com/Apollo-Lab-Yale/drift\_dagger}} 
\end{itemize}

%% file: 2-background.tex
\section{Background}

\subsection{Diffusion Policy Primer}

A denoising diffusion probabilistic model (DDPM) \cite{ho2020denoising, song2020denoising} consists of a forward process and a reverse process. In the forward process, Gaussian noise is gradually added to the training data, $x_0 \sim p(x_0)$, over $T$ discrete time steps. This process is governed by a predefined noise schedule, $\beta_t$, which controls how much noise is added at each step. Mathematically, the forward process is defined as:
\begin{align*}
&q(x_{1:T} \mid x_0) := \prod_{t=1}^T q(x_t \mid x_{t-1}), \\
&q(x_t \mid x_{t-1}) := \mathcal{N}(x_t; \sqrt{1 - \beta_t} x_{t-1}, \beta_t I),
\end{align*}
where $q(x_t \mid x_{t-1})$ is a Gaussian distribution with a mean of $\sqrt{1 - \beta_t} x_{t-1}$ and variance $\beta_t$. Intuitively, this step progressively adds noise to $x_0$, such that by the final step $x_T$, the data is almost entirely noise.

The reverse process aims to undo this noise, step by step, to recover the original data $x_0$. This is parameterized by a neural network, $\pi_\theta$, which predicts the noise added to $x_t$ at each step $t$. Using this prediction, the reverse process reconstructs the data from the noisy input $x_t$:
\begin{equation*}
x_{t-1} \sim p_\theta(x_{t-1} \mid x_t) := \mathcal{N}(x_{t-1}; \mu_k(x_t, \pi_{\theta}(x_t, t)), \sigma_t^2 I),
\end{equation*}
where $\mu_k(\cdot)$ computes the mean for the denoised data at step $t-1$, and $\sigma_t^2$ is a fixed variance term.

In the context of robotics, a diffusion policy \cite{chi_dp} adapts the DDPM framework for visuomotor control by treating robot actions as $x$ and conditioning the denoising process on robot observations, such as camera images or sensor data. Specifically, the noise prediction network $\pi_\theta$ takes the current noisy action representation $x_t$ and the observations as inputs and predicts the noise to remove. Architectures like U-Nets \cite{ronneberger2015u} or transformers \cite{vaswani2017attention} are commonly used for $\pi_\theta$. Diffusion policies are typically trained offline using BC, where the model learns to mimic expert demonstrations.

\subsection{Ranks in Diffusion Models}
\label{sec:ranks_in_diffusion_models}
The rank of a matrix, defined as the maximum number of linearly independent rows or columns \cite{strang2022introduction}, is closely tied to the expressiveness and representational power of a model. For example, in linear models, the rank of the weight matrix determines the dimensionality of the feature space that the model can effectively capture. Weight matrices with low ranks often correspond to models with limited capacity but faster training, while those with high ranks indicate overparameterization, which can improve optimization but at the cost of slower training \cite{du2018power}.

In the context of a diffusion policy that employs a U-Net with one-dimensional convolutional blocks as its network backbone, a weight matrix for each convolutional block \( W \in \mathbb{R}^{m \times n} \) can be created by reshaping a corresponding weight tensor $W_{\text{conv}} \in \mathbb{R}^{C_{\text{out}} \times C_{\text{in}} \times k}$ via
$$
W = \texttt{reshape}(W_{\text{conv}}, (m, n)),
$$ 
where \( C_{\text{out}} \) is the number of output channels, \( C_{\text{in}} \) is the number of input channels, and \( k \) is the kernel size. The reshaping can be performed by setting $m=C_{\text{out}} * k$ and $n=C_{\text{in}}$ or other equivalent view transformations.

The highest possible rank \( r_{\text{max}} \) of this weight matrix is bounded by:
$$
r_{\text{max}} \leq \min(m, n).
$$

\subsection{Problem Statement}
\label{sec:problem_statement}

In this work, we investigate diffusion policies that use a U-Net backbone composed of one-dimensional convolutional blocks, as introduced above.  For each convolutional block with weight \( W \) and highest possible rank \( r_{\text{max}} \) in a diffusion policy \( \pi_{\theta} \), we aim to enable dynamic adjustment of the rank \( r \) of a trainable segment of the weight matrix, \( W_{\text{train}} \), for any $r$ integer satisfying \( 1 \leq r \leq r_{\text{max}} \).  We assume all weight matrices $W$ throughout the network $\pi_\theta$ will vary uniformly based on $r$.  Importantly, $r$ should remain adjustable throughout the learning process without introducing instability or computational overhead.  



%% file: 3-related_works.tex
\section{Related Works}
\label{sec:related_works}

\subsection{Overparameterization and Intrinsic Rank}
\label{sec:overparam_and_rank}
Overparameterization, where models have more parameters than necessary to fit the training data, is a key factor behind the success of modern machine learning \cite{krizhevsky2012imagenet, kaplan2020scaling}. Large models such as diffusion models \cite{ho2020denoising} and transformers \cite{vaswani2017attention} excel in tasks like image synthesis \cite{rombach2022high}, robotic manipulation \cite{chi_dp}, and language generation \cite{zhao2023survey}. Although overparameterized models offer impressive performance, their size also poses significant challenges for training and fine-tuning due to high computational and memory requirements.

To tackle these challenges, researchers have observed that overparameterized models often reside in a low-dimensional subspace \cite{aghajanyan_intrinsic_dim, li_intrinsic_dim, xu2019trained}. Pre-deep
learning approaches like dynamical low-rank approximation \cite{koch2007dynamical} assume a derivable target matrix, unsuitable for deep learning where the target weight matrix is unknown. More modern techniques like Low-Rank Adaptation (LoRA) \cite{hu_lora, dettmers_qlora} fine-tune a small low-rank adapter while keeping the main model frozen. Although LoRA and its
variants like DyLoRA~\cite{valipour2022dylora} and QLoRA~\cite{dettmers_qlora} effectively reduces computational costs, it is primarily suited for fine-tuning pre-trained models and is less practical for training models from scratch due to its fixed low-rank structure \cite{liu2024dora}.

For LoRA, the need to merge and re-inject adapters during rank adjustments and the increased parameters during forward pass can destabilize training and increase computational overhead. In contrast, the DRIFT framework dynamically adjusts trainable ranks without adding new parameters or introducing instability. By using SVD to partition weight matrices into trainable and frozen components, DRIFT maintains stability and efficiency, making it well-suited for training overparameterized diffusion policies from scratch.

\subsection{Imitation Learning and Diffusion Policy}
\label{sec:il_dp}

Imitation Learning (IL) is a widely studied policy learning paradigm applied to various robotic scenarios. IL involves collecting demonstration data from an expert and training a neural network using supervised learning techniques \cite{zare2024survey}. Before the emergence of diffusion policies, IL research focused on improving sample efficiency and mitigating compounding errors through strategic sample collection and labeling \cite{spencer2021feedback}. \citet{ross2011reduction} first address these challenges with an iterative, interactive IL method. This approach collects additional demonstration rollouts using a suboptimal policy and refines the trajectories with corrections provided by an expert. Building on this work, expert-gated \cite{kelly2019hg, sun_mega_dagger} and learner-gated \cite{hoque2021thriftydagger, hoque2021lazydagger} methods allow experts or learners to dynamically take or hand over control during online rollout collection, which further improves data efficiency.

These methods primarily rely on interactive demonstration strategies and typically utilize simple neural network architectures, such as Multi-Layer Perceptrons (MLPs) \cite{jin2020geometric, loquercio2021learning, zhou2024developing} or Long Short-Term Memory (LSTM) networks \cite{cai2019vision, huang2020real, wu2024deep}. Historically, these interactive IL methods often employ shallow and small MLPs or LSTM, which are constrained by their relatively small number of parameters, limiting the performance of the learned policies.

Diffusion policies \cite{chi_dp} shift the focus of IL research to leveraging the representational power of overparameterized models. Inspired by generative models \cite{ho2020denoising, song2020denoising}, diffusion policies use large networks to achieve strong performance in various tasks. However, the computational demands of these models create challenges for both training and inference. Few existing works attempt to integrate interactive IL with diffusion models \cite{lee2024diff, zhang2024diffusion}. For example, \citet{lee2024diff} leverage diffusion loss to better handle multimodality; however, this work focuses on a robot-gated interactive approach rather than an expert-gated one and does not contain real-world experiments. Similarly, while \citet{zhang2024diffusion} employ diffusion as a policy representation, the primary innovation in this work lies in using diffusion for data augmentation rather than improving the interactive learning process. Notably, neither approach addresses the critical issue of reducing batch training time, which is essential for making online interactive learning with large models more practical and usable. Recent efforts to accelerate diffusion policies focus on inference through techniques like distillation \cite{prasad2024consistency, wang2024one}, but no existing work focuses on improving training efficiency. As a result, diffusion policy research remains largely confined to offline training scenarios \cite{sridhar2024nomad, sun2024comparative, ze20243d}. 

%% file: 4-drift_framework.tex
\section{DRIFT Framework}
\label{sec:drift_framework}

\subsection{Overview}
The DRIFT framework is designed to dynamically adjust trainable ranks in a diffusion policy, allowing the number of trainable parameters to change throughout the training process. This flexibility enables efficient training from scratch by leveraging the intrinsic low-rank structure of overparameterized models. As covered in \S\ref{sec:problem_statement}, the rank adjustment process must ensure training stability and avoid introducing additional computational overhead. These considerations are critical for training from scratch but are often overlooked by existing methods like LoRA \cite{hu_lora}, which are primarily designed for fine-tuning. Applying methods like LoRA for dynamic rank adjustment during training from scratch can result in higher computational time for forward pass and instability due to the need for merging and re-injecting newly initialized low-rank adapters, which disrupts the training process.

To achieve dynamic trainable rank adjustment while maintaining training stability, we propose \textit{rank modulation}. Rank modulation uses the Singular Value Decomposition (SVD) \cite{strang2022introduction} to partition ranks into trainable and frozen sections. This approach avoids introducing new parameters, ensures that computational costs for the backward pass decrease as the trainable ranks are reduced, and maintains a constant computational cost for the forward pass.

In addition to rank modulation, the framework can incorporate a \textit{rank scheduler} to coordinate the dynamic adjustment of trainable ranks. While rank modulation facilitates the adjustment itself, the rank scheduler determines how the trainable ranks may automatically evolve during training. The rank scheduler uses a decay function that calculates the current number of trainable ranks based on the training epoch, the maximum rank, and the desired terminal rank of the policy.


The rank modulation and rank scheduler components can work together to enable efficient training of diffusion policies by dynamically balancing representational power and computational efficiency.  


\subsection{Rank Modulation}
\label{sec:rank_modulation}

Rank modulation takes inspiration from LoRA \cite{hu_lora}, which employs an adapter with small trainable ranks during fine-tuning. LoRA achieves this by injecting additional low-rank weight matrices into the network layers, allowing only these matrices to be updated during backpropagation. For the one-dimensional convolution blocks in a U-Net architecture used by diffusion models, LoRA would replace the original convolutional layer with:
\[
\text{Conv}_{\text{LoRA}}(x) = W_{\text{conv}} \circledast x + \alpha((W_{\text{up}} \times W_{\text{down}}) \circledast x),
\]
where $\circledast$ denotes convolution, $W_{\text{conv}}$ is the original convolution weight tensor of shape $(C_{\text{out}}, C_{\text{in}}, k)$, with $C_{\text{out}}$ and $C_{\text{in}}$ denoting the output and input channels, and $k$ is the kernel size. Two low-rank matrices, $W_{\text{down}} \in \mathbb{R}^{r \times C_{\text{in}} \times k}$ and $W_{\text{up}} \in \mathbb{R}^{C_{\text{out}} \times r \times k}$, are introduced, where $r \ll C_{\text{in}}$. A scaling factor $\alpha$ further controls the magnitude of the low-rank update. During backpropagation, gradients are computed only for $W_{\text{down}}$ and $W_{\text{up}}$, thus lowering the number of trainable ranks.

Despite these benefits, LoRA has several limitations when applied to IL that trained from scratch. Since LoRA is essentially an approximation of the full-rank weight, it relies on the main weight $W_{\text{conv}}$ being thoroughly pre-trained. Otherwise, the low-rank approximation may limit the representational power of the model and prevent it from fully benefiting from overparameterization. Furthermore, injecting LoRA blocks adds complexity to the forward pass. Due to the merging of $W_{\text{up}}$ and $W_{\text{down}}$, the additional convolution with LoRA blocks results in a time complexity of $O(C_{\text{out}} \times C_{\text{in}} \times r \times k)$, which increases computational overhead proportional to the rank $r$ compared to the time complexity of the original convolution $O(C_{\text{out}} \times C_{\text{in}} \times k)$. While this computational overhead is often negligible when fine-tuning a pre-trained model with a rank of less than 4 \cite{hu_lora}, training a model from scratch requires low-rank adapters with significantly higher number of trainable ranks to effectively leverage both overparameterization and low-rank structure (as we will demonstrate in \S\ref{sec:abl_tr}). This increase in trainable ranks amplifies the computational cost associated with $r$, making it a critical consideration in such scenarios.

Finally, if LoRA is used with a dynamic rank scheduler (discussed in \S\ref{sec:rank_scheduler}), new LoRA blocks must be injected each time the rank changes, introducing freshly initialized parameters and destabilizing training. Consequently, repeatedly merging and reinjecting LoRA blocks is inefficient when the trainable rank is adjusted on the fly. 



To address these limitations, we propose rank modulation, which leverages an SVD structure to decompose weight matrices into components designated as either frozen or trainable ranks. More specifically, consider a weight matrix 
$
  W \in \mathbb{R}^{m \times n}
$
that can be created from a corresponding weight tensor \( W_{\text{conv}} \in \mathbb{R}^{C_{\text{out}} \times C_{\text{in}} \times k}\) for a one-dimensional convolutional layer by reshaping, e.g., $C_{\text{out}} * k$ becomes $m$ and $C_{\text{in}}$ becomes $n$ (as covered in \S\ref{sec:ranks_in_diffusion_models}).  This matrix can be refactored into three matrices using the SVD:  

$$
   W = U\,\Sigma\,V^T,
$$
where 
$
  U \in \mathbb{R}^{m \times m},\,
  \Sigma \in \mathbb{R}^{m \times n},\,
  V \in \mathbb{R}^{n \times n}.
$ 
$U$ and $V$ are orthonormal matrices that represent rotations or reflections, while $\Sigma$ is a diagonal matrix containing scaling factors.
We can further split $U, \Sigma,$ and $V$ at a specified rank $r$ to partition trainable and frozen part of each matrix:
\begin{align*}
U &= 
\bigl[\,
   U_{\text{train}}\ \ U_{\text{frozen}}
\bigr],
\\
\Sigma &= 
\begin{bmatrix}
  \Sigma_{\text{train}} & 0_{r \times (n - r)} \\
  0_{(m - r) \times r} & \Sigma_{\text{frozen}}
\end{bmatrix},
\\
V &= 
\bigl[\,
   V_{\text{train}}\ \ V_{\text{frozen}}
\bigr].
\end{align*}
Accordingly, we represent trainable weight $W_{\text{train}}$ and frozen weight $W_{\text{frozen}}$ as:
\begin{align*}
W_{\text{train}} &= U_{\text{train}} \,\Sigma_{\text{train}}\,V_{\text{train}}^T, \\
W_{\text{frozen}} &= U_{\text{frozen}} \,\Sigma_{\text{frozen}}\,V_{\text{frozen}}^T,
\end{align*}
where $\Sigma_{\text{frozen}}$ holds smaller singular values than $\Sigma_{\text{train}}$.
During training, \(\{U_{\text{train}}, \Sigma_{\text{train}}, V_{\text{train}}\}\) are the only parameters that receive gradient updates (rank-\(r\) subspace), while \(\{U_{\text{frozen}}, \Sigma_{\text{frozen}}, V_{\text{frozen}}\}\) remain fixed. 


The procedure described above, converting $W_{\text{conv}}$ into $W_{\text{train}}$ and $W_{\text{frozen}}$ using SVD and partitioning, is performed at the start of each epoch if the number of trainable ranks has changed since the previous epoch. This procedure also reorthonormalizes the $U$ and $V^\top$ matrices. While it is possible to reorthonormalize the $U$ and $V^\top$ matrices more frequently (such as after each gradient update) using QR decomposition, preliminary experiments showed that this introduced significant computational overhead without measurable performance gains (see Appendix \S\ref{sec:appendix_additional_comp}). Therefore, our implementation defaults to reorthonormalizing only via full SVD at the start of an epoch whenever the number of trainable ranks changes. 



Unlike LoRA, rank modulation performs a single convolution using the full \(W = W_{\text{train}} + W_{\text{frozen}}\) via another simple view transformation in memory:
$$
W_{\text{conv}} = \texttt{reshape}(W, (C_{\text{out}}, C_{\text{in}}, k)). 
$$
Hence, the forward time complexity remains the same as a standard convolution. Because no additional new parameters are introduced during the training process, rank modulation can also preserve stable updates even when the number of trainable ranks changes.

\subsection{Rank Scheduler}

Building on rank modulation, which dynamically adjusts the number of trainable ranks while maintaining stable training, we introduce a rank scheduler to further exploit the low-rank structure. The rank scheduler, inspired by the noise scheduler in diffusion models that dynamically adjusts the added noise \cite{ho2020denoising}, is designed to improve training efficiency without compromising performance.

The rank scheduler uses a decay function to compute the current number of trainable ranks $r_i$ based on the current training epoch index $i$, and the maximum and minimum trainable ranks, $r_{\text{max}}$ and $r_{\text{min}}$. Once $r_i$ is determined, the trainable ranks are adjusted depending on the low-rank adapters. For instance, with LoRA, this process involves merging the current LoRA blocks and reinstantiating new blocks with the updated trainable ranks. In the case of rank modulation, this process entails computing the SVD of $W$ to produce updated $W_{\text{train}}$ and $W_{\text{frozen}}$ matrices.

In this work, we implement and evaluate four decay functions, which are linear, cosine, sigmoid, and exponential:
$$
r_{\text{linear}} = \left\lfloor r_{\text{max}} - (r_{\text{max}} - r_{\text{min}}) \times \left(\frac{i}{T}\right) \right\rfloor
$$
$$
r_{\text{cosine}} = \left\lfloor r_{\text{min}} + 0.5 \times (r_{\text{max}} - r_{\text{min}}) \times \left(1 + \cos\left(\pi \times \frac{i}{T}\right)\right) \right\rfloor
$$
$$
r_{\text{sig}} = \left\lfloor r_{\text{max}} - \frac{(r_{\text{max}} - r_{\text{min}})}{1 + e^{-\tau \times (i - t_{m})}} \right\rfloor 
$$
$$
r_{\text{exp}} = \left\lfloor r_{\text{min}} + (r_{\text{max}} - r_{\text{min}}) \times e^{-\tau \times i} \right\rfloor
$$
where $i$, $T$, and $t_m$ are the current, total, and midpoint of the number of training epochs, respectively, $\tau$ denotes steepness, and $\lfloor\cdot\rfloor$ is the floor function. 

\label{sec:rank_scheduler}

%% file: 5-drift_dagger.tex
\section{DRIFT-DAgger}
\label{sec:drift_dagger}

DRIFT-DAgger combines the sample efficiency of interactive IL with the computational efficiency of low-rank training methods, making it well-suited for training large policies interactively.

Algorithm \ref{alg:cap} outlines the DRIFT-DAgger procedure. Similar to previous interactive IL methods, DRIFT-DAgger consists of an offline bootstrapping stage followed by an online adaptation stage. The process begins with an initial policy $\pi_{N_0}$, parameterized by a neural network that can adjust the number of trainable ranks. Although DRIFT-DAgger is proposed as an instantiation of the DRIFT framework, the adjustment of trainable ranks can be achieved through any kind of low-rank adapters other than the rank modulation proposed in \S\ref{sec:rank_modulation}, such as LoRA.

In the offline bootstrapping stage, DRIFT-DAgger trains the policy $\pi_{N_i}$ on an offline dataset $\mathcal{D}_B$ over several epochs $i$ using BC, similar to prior interactive IL methods. However, unlike these methods, DRIFT-DAgger optionally employs a rank scheduler that gradually reduces the number of trainable ranks during training. The rank scheduler uses a decay function based on the epoch index $i$, along with the highest possible ranks ($r_{\text{max}}$) and terminal trainable ranks ($r_{\text{min}}$) for the policy network. This approach reduces computational costs while maintaining performance. Details of the rank scheduler are presented in \S\ref{sec:rank_scheduler}.

If the rank scheduler is not used, the number of trainable ranks is set fixed at $r_{min}$ after offline bootstrapping and before transitioning to the online adaptation stage. At the end of offline bootstrapping, the offline dataset $\mathcal{D}_B$ is merged into a global dataset $\mathcal{D}$ for further use in online adaptation.

During the online adaptation stage, the learner policy interacts with the environment through rollouts. At each iteration $j$, the learner executes a rollout in the environment. If the expert policy $\pi_{exp}$ detects that the learner has deviated from the desired trajectory, the expert intervenes, taking control to provide corrective demonstrations. The expert can be a human teleoperator, an algorithm, or another neural network. These demonstrations are recorded in a dataset specific to the current rollout $\mathcal{D}_j$. After each rollout, $\mathcal{D}_j$ is merged into the global dataset $\mathcal{D}$, and the learner policy $\pi_{N_{\mathcal{I}+j}}$ is updated using the expanded dataset $\mathcal{D}$.

The full procedure of DRIFT-DAgger leverages low-rank training and online interaction to achieve better sample and training efficiency.

\begin{algorithm}
\caption{DRIFT-DAgger}\label{alg:cap}
\SetKwProg{Procedure}{procedure}{}{}
\Procedure{\textnormal{DRIFT-DAgger}$(\pi_{exp}$, $\pi_{N_0}$, $\mathcal{D}_B)$}{}
\For{\textnormal{offline epoch} $i = 1, 2, \cdots ,\mathcal{I}$}{
    train $\pi_{N_i}$ on offline dataset $\mathcal{D}_B$\\
    \If{\textnormal{use rank scheduler}}{
       $r_{i} = \texttt{Decay Function}(i, r_{\text{min}}, r_{\text{max}})$\\
       $\pi_{N_i} = \texttt{Rank Reduction}(r_{i}, \pi_{N_i})$\\
    }

\If{\textnormal{not use rank scheduler}}{
    $\pi_{N_{\mathcal{I}}} = \texttt{Rank Reduction}(r_{\text{min}}, \pi_{N_{\mathcal{I}}})$
}
}

$\mathcal{D} \gets \mathcal{D}_B$\\

\For{\textnormal{online iteration} $j = 1, 2, \cdots ,\mathcal{J}$}{
        \For{\textnormal{timestep} $t \in T$ \textnormal{of online rollout} $j$}{
            \If{$\pi_{exp}$ \textnormal{takes control}}{
                $observation \gets \texttt{rollout}_{j}^t$\\
                $action \gets \pi_{exp}(observation)$\\
                $\mathcal{D}_j \gets (observation, action)$\\
            }
        }
        $\mathcal{D} \gets \mathcal{D} \cup \mathcal{D}_j$
        
        Train $\pi_{N_{\mathcal{I}+j}}$ on $\mathcal{D}$
    }
\Return $\pi_{N_{{\mathcal{I}}+\mathcal{J}}}$
\end{algorithm}

%% file: 6-simulation_evaluation.tex
\section{Simulation Evaluation}
\label{sec:simulation_evaluation}

We evaluate the proposed DRIFT framework, instantiated in the DRIFT-DAgger algorithm, through extensive simulation experiments and ablation studies. All experiments are conducted using the PyTorch framework \cite{pytorch}, with the UNet-based diffusion policies that enable RGB perception following the specifications from \citet{chi_dp}. The batch size and learning rate is set to 256 and $10^{-4}$ for all experiments. We use Adam \cite{kingma2014adam} as the optimizer. Training is performed on a desktop PC with an AMD PRO 5975WX CPU, 4090 GPU, and 128GB RAM. To ensure a fair comparison with interactive methods like HG-DAgger~\cite{kelly_hg_dagger} and DRIFT-DAgger, we implement BC with an incremental dataset during the online phase, similar to the interactive loop of HG-DAgger and DRIFT-DAgger.

\subsection{Decay Functions}
\label{sec:abl_rs}

\begin{figure}[t]
\centering
\begin{minipage}{\columnwidth}
\includegraphics[width=\columnwidth]{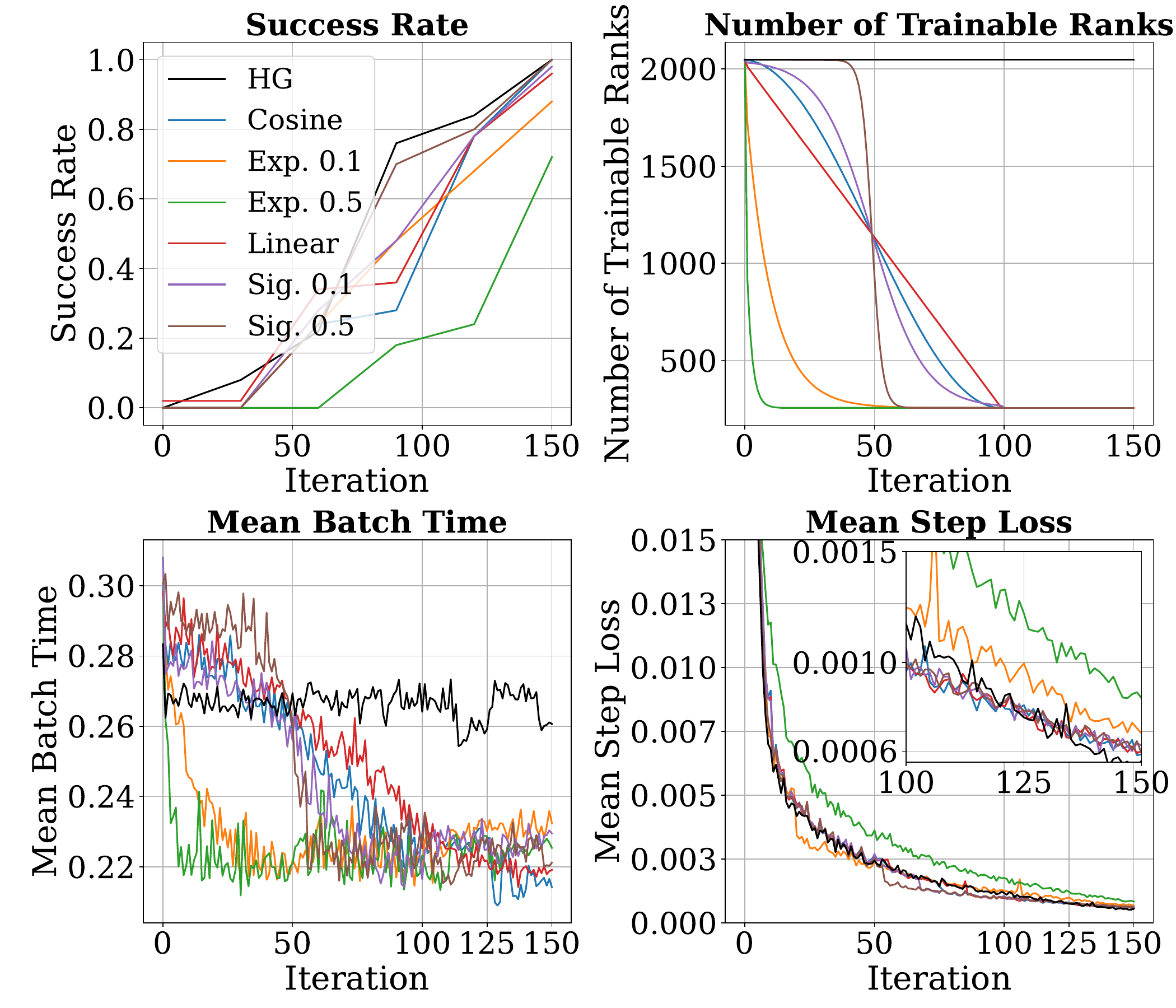}
\caption{Experimental results of DRIFT-DAgger with different decay functions for the rank scheduler. We use HG-DAgger (HG) as a baseline for comparison.}
\label{fig:decay_functions}
\end{minipage}

\vspace{1em} 

\begin{minipage}{\columnwidth}
    \centering
    \renewcommand\arraystretch{1.2}
    \captionof{table}{Summary of experimental results on mean batch training time (MBT) and success rate with different rank decay functions for DRIFT-DAgger. \label{tab:training_scheduler}}
    \resizebox{\columnwidth}{!}{ 
    \begin{tabular}{l|c|c|c|c}
    \hline 
    Function & \makecell{Success \\ Rate} & \makecell{MBT \\ (Offline)} & \makecell{MBT \\ (Online)} & \makecell{MBT \\ (All Stages)} \\ 
    \hline
    HG                  & $\mathbf{1.0}$ & 0.27 & 0.27 & 0.27 \\
    Linear              & $0.96$ & $0.26$ & $0.22$ & $0.25$ \\
    Cosine              & $\mathbf{1.0}$ & $0.26$ & $0.22$ & $0.25$ \\
    Exp. 0.1            & $0.88$ & $0.23$ & $0.23$ & $0.23$ \\
    Exp. 0.5            & $0.72$ & $\mathbf{0.22}$ & $\mathbf{0.22}$ & $\mathbf{0.22}$ \\
    Sig. 0.1            & $0.98$ & $0.25$ & $0.23$ & $0.24$ \\
    Sig. 0.5            & $\mathbf{1.0}$ & $0.26$ & $0.22$ & $0.24$ \\
    \hline
    \end{tabular}
    }
\end{minipage}
\end{figure}
To identify the scheduling strategy that best exploits the benefits of reduced-rank training while balancing the trade-off between training time and policy performance, we evaluate six variants of four decay functions for the rank scheduler. These functions dynamically adjust the number of trainable ranks as training progresses. The decay functions considered include linear, cosine, exponential, and sigmoid, as covered in \S\ref{sec:rank_scheduler}, with steepness parameters $\tau$ set to 0.1 and 0.5 for exponential and sigmoid. 


We use DRIFT-DAgger with rank modulation and rank scheduler for this ablation study. We set $r_{\text{min}}$ to 256 for all DRIFT-DAgger variants. Since both BC and HG-DAgger employ full-rank training, they should exhibit similar batch training times. Given that HG-DAgger outperforms BC in terms of sample efficiency\cite{kelly2019hg}, we use HG-DAgger as the baseline for full-rank training methods. 

To assess performance and training efficiency across different decay functions, we evaluate the policy's success rate, with higher values indicating better task completion. We also track the step loss during training as a measure of convergence. To assess training efficiency, we record the mean batch training time per epoch, separately for the offline stage, online stage, and their combination, interpreting it alongside the success rate and step loss.


The ablation study is conducted on a pick-and-place scenario from the Manipulation with Viewpoint Selection (MVS) tasks \cite{sun2024comparative}. The pick-and-place scenario, as illustrated in Fig. \ref{fig:simulation_environments}, requires the agent to pick up a green cube and place it in a red region.  All MVS tasks involve two UFactory xArm7 robots\footnote{\href{https://www.ufactory.us/product/ufactory-xarm-7}{https://www.ufactory.us/product/ufactory-xarm-7}} mounted on linear motors, where one arm has a gripper and the other is equipped with a camera. Mounted on one end effector, the camera enables active perception, working in synergy with the gripper on the other end effector to cooperatively execute the manipulation task. The state-action space for all MVS tasks is a reduced end-effector space for the dual-arm system, with automatically computed camera orientation using an additional Inverse Kinematics (IK) objective.

The learning process uses 100 offline demonstration rollouts, 100 offline bootstrapping epochs, and 50 online iterations. We plot the experimental results by combining the number of offline epochs and online iterations, resulting in a total of 150 iterations.

The results of this experiment are presented in Fig. \ref{fig:decay_functions} and summarized in Table \ref{tab:training_scheduler}. While all decay functions reduce batch training time compared to full-rank training represented by HG-DAgger, some decay functions lead to decreased policy performance, as indicated by the success rates in Fig. \ref{fig:decay_functions} and Table \ref{tab:training_scheduler}. Notably, the exponential decay functions, due to their aggressive reduction of trainable ranks, underperform relative to the other variants, despite yielding the lowest mean batch training time.

The linear decay function, while offering near-perfect policy performance, results in the highest training time among all variants, suggesting that it is less effective than the sigmoid decay functions in balancing training time with performance. We observe that the sigmoid functions, particularly with a steep decay parameter $\tau = 0.5$, strike the best balance between training time and policy performance. These functions maintain a high number of trainable ranks during the early training phase, allowing overparameterization to effectively minimize approximation error, as shown in the loss plot in Fig. \ref{fig:decay_functions}. This facilitates more efficient learning of the predominant behavior during the early stage of training, while still preserving the flexibility required for online adaptation.

\subsection{Terminal Rank}
\label{sec:abl_tr}


\begin{figure}[tbp]
\centering
\begin{minipage}{\columnwidth}
\includegraphics[width=\columnwidth]{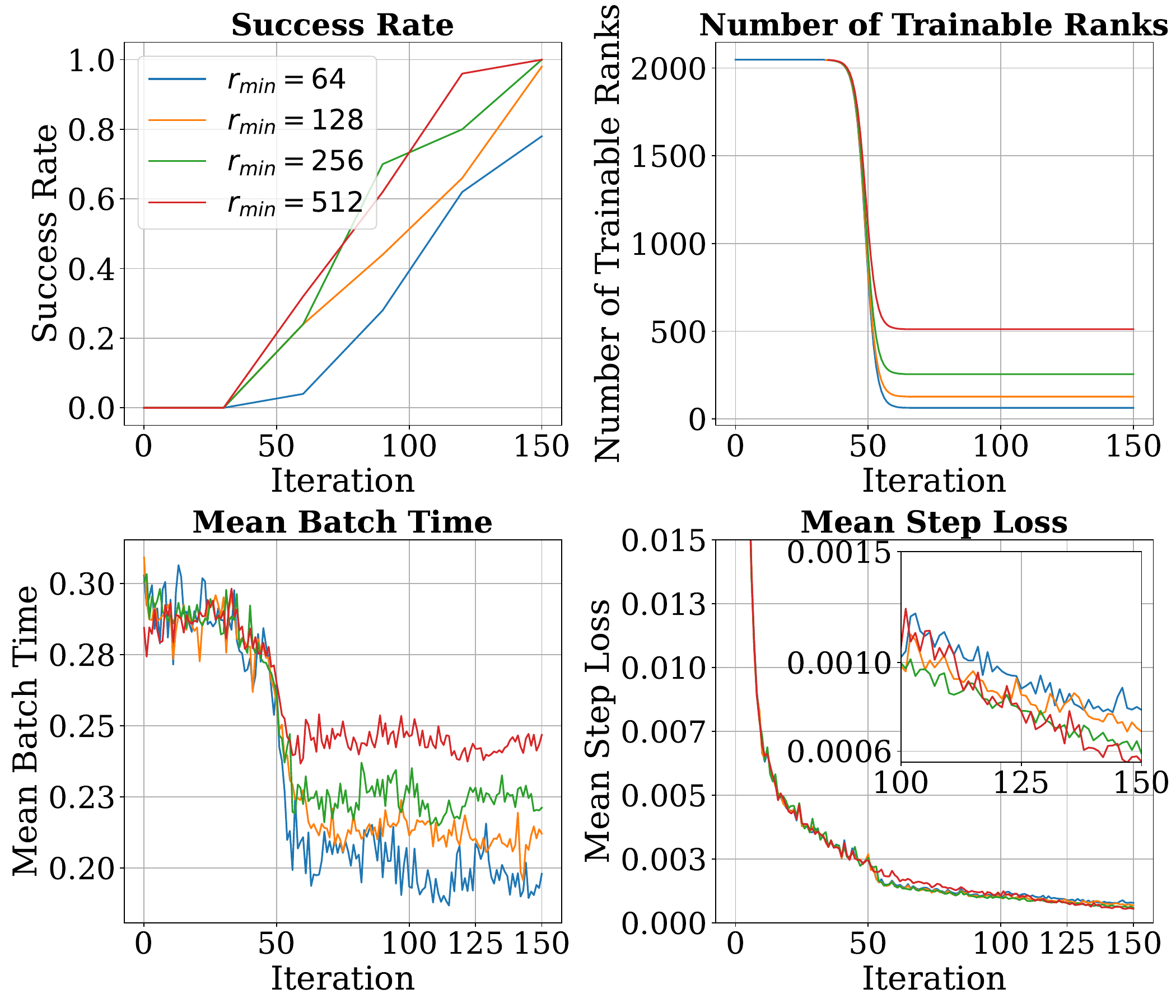}
\caption{Experimental results of DRIFT-DAgger with different terminal ranks $r_{\text{min}}$.}
\label{fig:terminal_ranks}
\end{minipage}

\vspace{1em} 

\begin{minipage}{\columnwidth}
    \centering
    \renewcommand\arraystretch{1.2}
    \captionof{table}{Summary of experimental results on mean batch training time (MBT) and success rate with different values for terminal ranks for DRIFT-DAgger. \label{tab:terminal_ranks}}
     
    \begin{tabular}{l|c|c|c|c}
    \hline 
    \makecell{$r_{\text{min}}$} & \makecell{Success \\ Rate} & \makecell{MBT \\ (Offline)} & \makecell{MBT \\ (Online)} & \makecell{MBT \\ (All Stages)} \\ 
    \hline
    64     & 0.78 & $\mathbf{0.24}$ & $\mathbf{0.19}$ & $\mathbf{0.22}$ \\
    128    & 0.98 & 0.25 & 0.21 & 0.23 \\
    256    & $\mathbf{1.0}$ & 0.26 & 0.22 & 0.24 \\
    512    & $\mathbf{1.0}$ & 0.27 & 0.24 & 0.26\\
    \hline
    \end{tabular}
\end{minipage}
\end{figure}

To explore the effect of different terminal ranks for the DRIFT framework, we conduct an ablation study by varying the terminal rank $r_{\text{min}}$ in DRIFT-DAgger with rank modulation and rank scheduler. We use the same experimental task, setup, and evaluation metrics as \S\ref{sec:abl_rs}, and use sigmoid decay function with $\tau$ set to 0.5 for rank scheduler. The terminal rank $r_{min}$ is set to 64, 128, 256, and 512 for comparison.

As shown in Fig. \ref{fig:terminal_ranks} and summarized in Table \ref{tab:terminal_ranks}, decreasing the terminal ranks reduces training time but also impacts policy performance, as reflected in the success rate. This performance degradation occurs due to the diminished representational power of the model when the number of trainable ranks is reduced. Compared to reduced-rank methods for fine-tuning, such as \cite{hu_lora}, where fine-tuning with an adapter requires only 4 trainable ranks, DRIFT-DAgger with reduced-rank training from scratch necessitates a significantly higher number of trainable ranks, given that there is a noticeable performance drop when $r_{\text{min}}$ is set to 64. This is to fully leverage the benefits of both better representation power from overparameterization and improved computational efficiency from intrinsic low ranks, as fine-tuning with extremely small adapters does not capture the full representational potential of the model.

Conversely, setting $r_{\text{min}}$ too high, such as 512, offers no clear benefit in terms of policy performance, as the model already achieves a perfect success rate. The approximation error, as indicated by the loss curve in Fig. \ref{fig:terminal_ranks}, also shows negligible differences between $r_{\text{min}}$ values of 256 and 512. We also run $r_{\text{min}}$ sweep for another MVS task, and the results follow similar trends as the pick-and-place scenario that $r_{\text{min}}=256$ provides the best balance of efficiency and performance (see Appendix \S\ref{sec:appendix_microwave_rmin_sweep}).

\begin{table}[ht!]
\centering
\renewcommand{\arraystretch}{1.2}
\setlength{\tabcolsep}{5pt}
\caption{The configurations of DRIFT-DAgger for all simulation and real-world tasks. For simulation, we use other neural network policies trained with BC as experts, and compare the cosine similarity of the expert action and learner action with the threshold to determine whether the expert take over or not.}
\normalsize
\label{tab:exp_config}
\begin{tabular}
{l|c|c|c|c|c}
    \hline
    & Task & 
    \makecell{Offline \\ Rollouts} & 
    \makecell{Boot. \\ Epochs}
    & 
    \makecell{Online \\ Iterations}
    & 
    \makecell{Cos. Sim. \\ Threshold}
    \\ 
    \hline
    \multirow{4}{*}{\rotatebox{90}{Sim.}} 
    & Robo-L      & 300  & 100  & 100 & 0.94  \\
    & Robo-C      & 275  & 100  & 100 & 0.95  \\
    & MVS-M       & 75   & 35   & 100 & 0.99  \\
    & MVS-PnP     & 100  & 100  & 100 & 0.99  \\
    \hline 
    \multirow{3}{*}{\rotatebox{90}{Real}} 
    & Real-BS     & 150  & 150  & 50  & - \\
    & Real-DA     & 200  & 200  & 50  & - \\
    & Real-DI     & 100  & 150  & 50  & - \\
    \hline
\end{tabular}
\end{table}

\subsection{Benchmark Comparison}
\label{sec:sim_exp}

\begin{figure*}[htb]
\centering
\begin{minipage}{\textwidth}
    \centering    
    \renewcommand{\arraystretch}{0.8} 
    \setlength{\tabcolsep}{0pt} 
    
    \begin{tabular}{cccc}
        \includegraphics[width=0.245\textwidth]{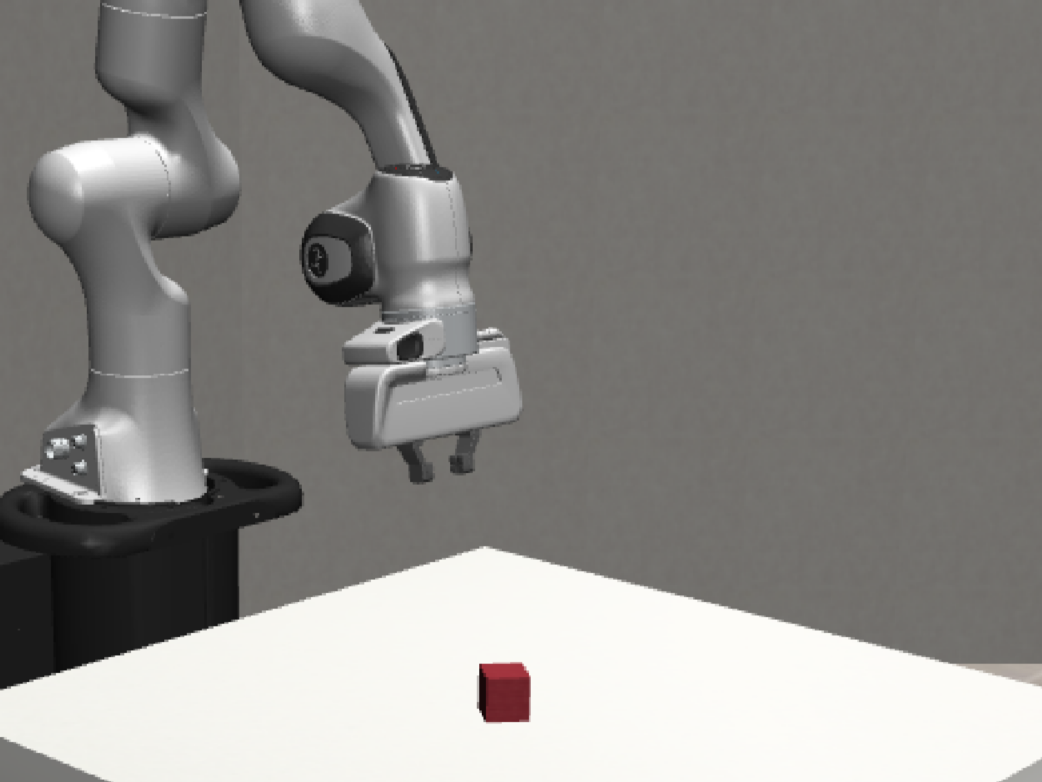} &
        \includegraphics[width=0.245\textwidth]{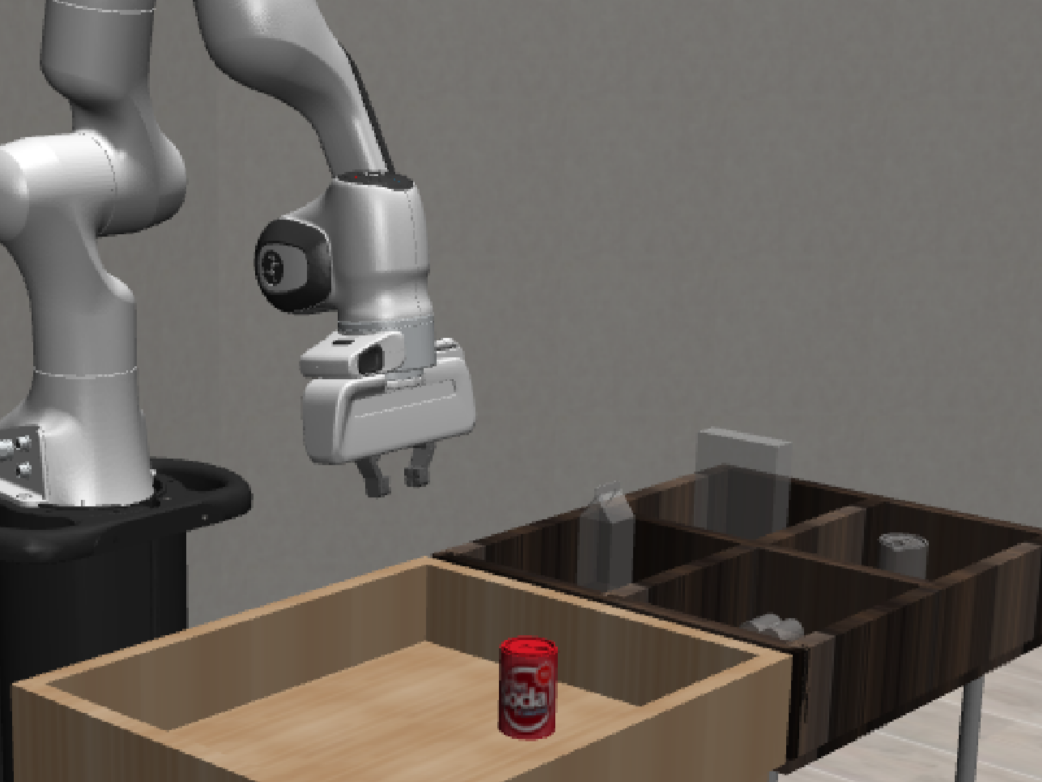} &
        \includegraphics[width=0.245\textwidth]{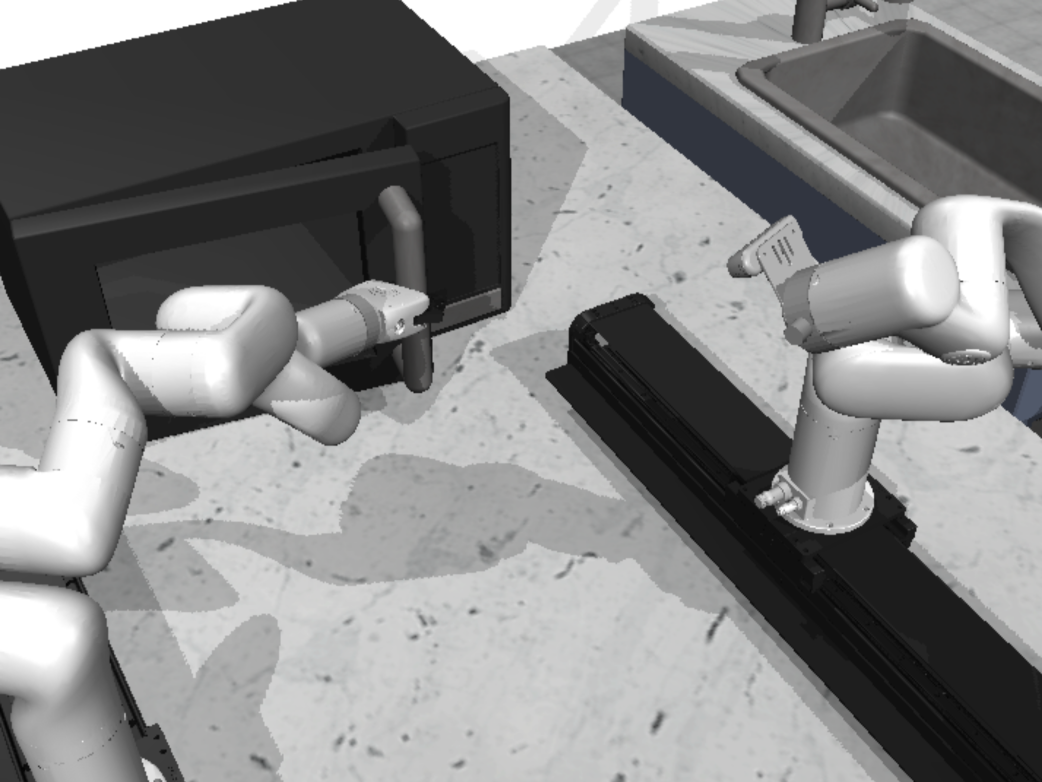} &
        \includegraphics[width=0.245\textwidth]{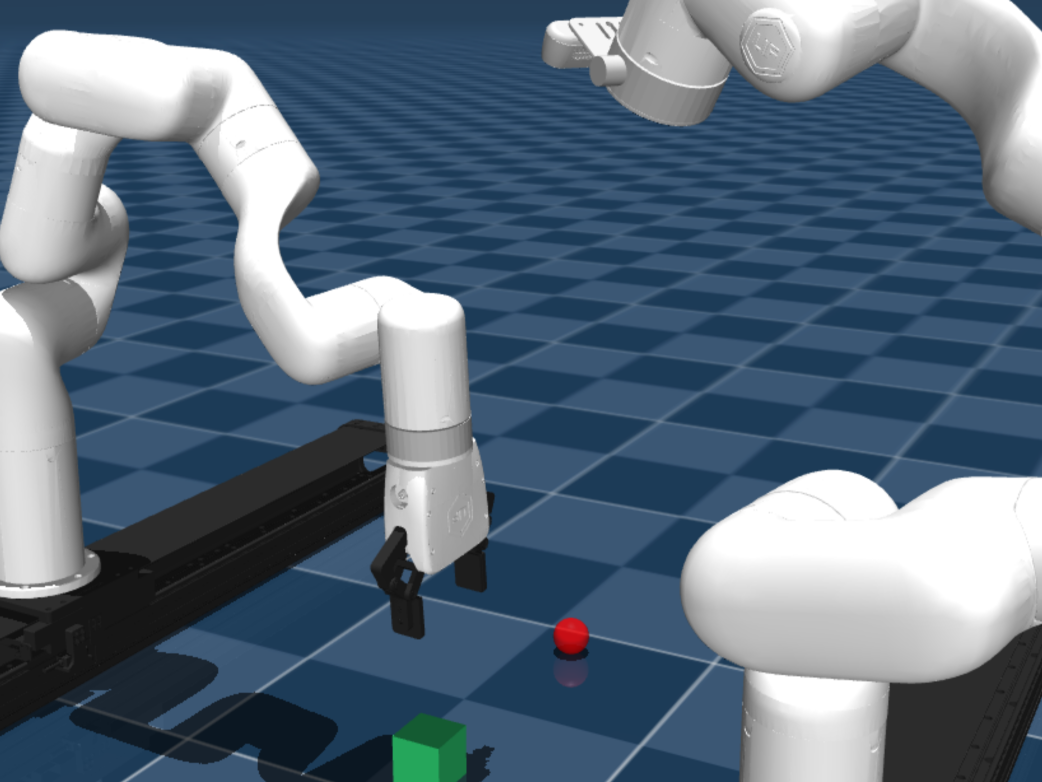} \\
        
        \begin{minipage}{0.25\textwidth}
            \centering
            \includegraphics[width=\textwidth]{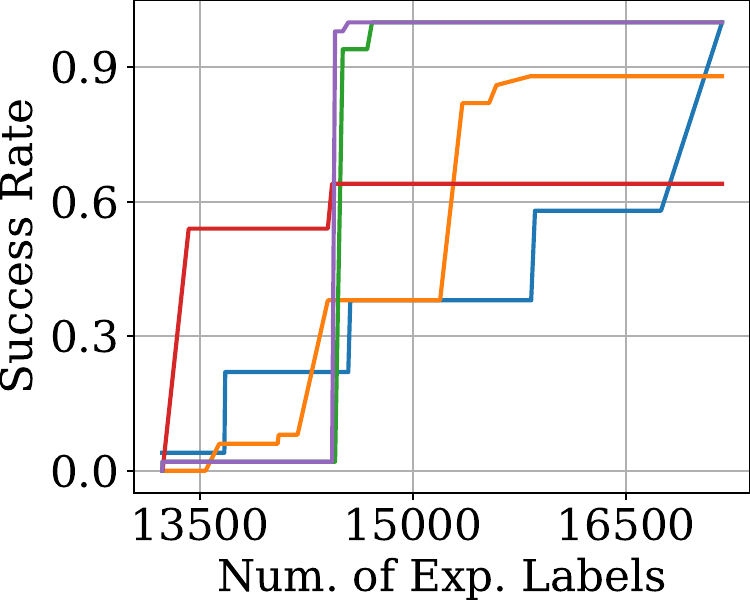}
            \subcaption{Robosuite - Lift}
        \end{minipage} &
        \begin{minipage}{0.25\textwidth}
            \centering
            \includegraphics[width=\textwidth]{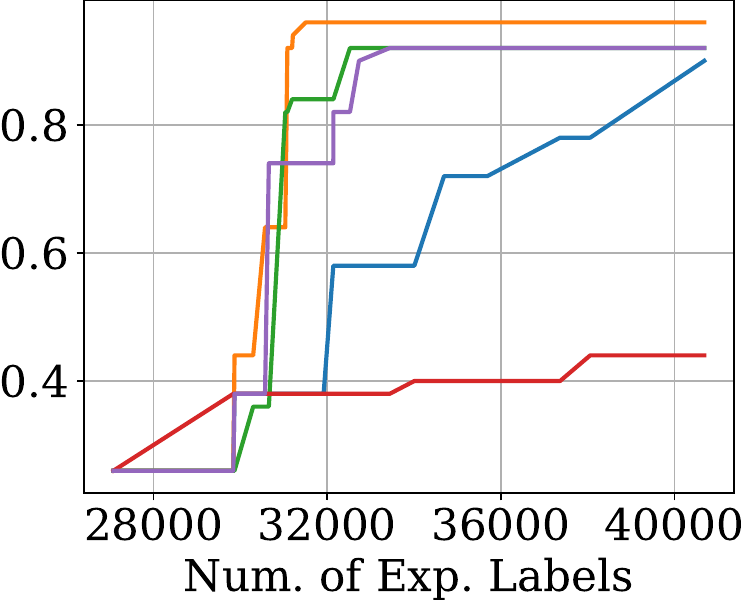}
            \subcaption{Robosuite - Can}
        \end{minipage} &
        \begin{minipage}{0.25\textwidth}
            \centering
            \includegraphics[width=\textwidth]{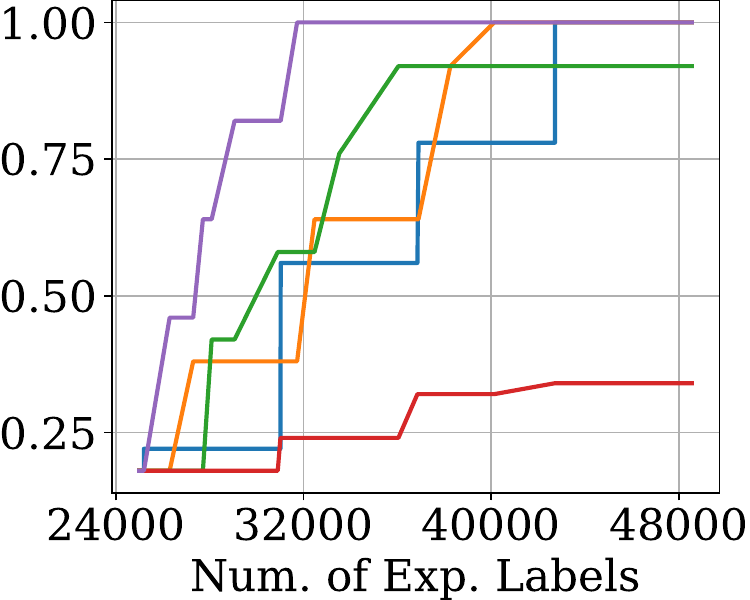}
            \subcaption{MVS - Microwave}
        \end{minipage} &
        \begin{minipage}{0.25\textwidth}
            \centering
            \includegraphics[width=\textwidth]{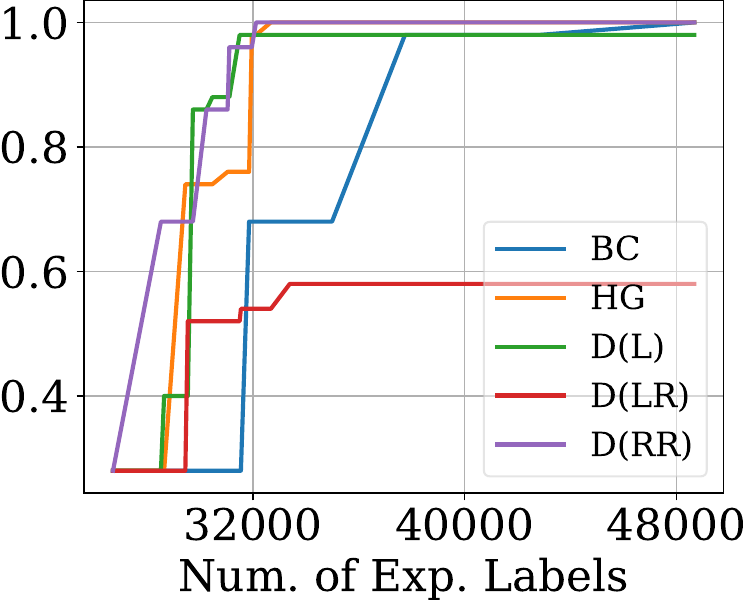}
            \subcaption{MVS - Pick and Place}
        \end{minipage} \\
    \end{tabular}
    \caption{The upper row shows the simulation scenarios from robosuite and Manipulation with Viewpoint Selection (MVS) tasks. The lower row shows the plots of success rate with respect to the number of expert labels. HG, D(L), D(LR), and D(RR) represent HG-DAgger, DRIFT-DAgger with LoRA adapters that are only instantiated with $r_{\text{min}}$ when switching to online mode, DRIFT-DAgger with LoRA and rank scheduler, and DRIFT-DAgger with rank modulation and rank scheduler.}
    \label{fig:simulation_environments}
\end{minipage}

\vspace{1em}

\begin{minipage}{\textwidth}
    \centering
    \renewcommand\arraystretch{1.2}
    \setlength{\tabcolsep}{1.1pt}
    \captionof{table}[Simulation Scenarios]{Summary of experimental results from simulation scenarios. The metrics include success rate (SR), mean and standard deviation of task duration (MSD), number of expert labels (NEL), and cumulative training time (CT). CT is measured in hours, MSD is measured in steps and at the scale of $\times 10^{2}$, and NEL is at the scale of $\times 10^{4}$}
    \label{tab:sim_scenarios}
    
    \begin{tabular}{l|cccc|cccc|cccc|cccc}
        \hline
        & \multicolumn{4}{c|}{Robosuite - Lift} & \multicolumn{4}{c|}{Robosuite - Can} &  \multicolumn{4}{c|}{MVS - Microwave} & \multicolumn{4}{c}{MVS - Pick and Place} \\
        \cline{2-17}
        & SR & MSD & NEL & CT & SR & MSD & NEL & CT & SR & MSD & NEL & CT & SR & MSD & NEL & CT\\
        \hline          
        Expert & $1.00$ & $0.43 \pm 0.03$ & - & - 
               & $0.98$ & $1.17 \pm 0.56$ & - & -  
               & $1.00$ & $3.00 \pm 0.29$ & - & - 
               & $0.92$ & $3.03 \pm 0.69$ & - & - \\

        BC & $\mathbf{1.00}$ & $0.58 \pm 0.61$ & $1.71$ & $1.66$
            & $0.90$ & $1.52 \pm 1.19$ & $4.06$ & $3.56$
            & $\mathbf{1.00}$ & $3.17 \pm 0.75$ & $4.85$ & $2.66$
            & $\mathbf{1.00}$ & $2.61 \pm 0.23$ & $4.86$ & $3.76$\\  
            
        HG & $0.88$ & $1.68 \pm 1.40$ & $1.58$ & $1.62$
           & $\mathbf{0.96}$ & $\mathbf{1.20 \pm 0.80}$ & $\mathbf{3.15}$ & $3.30$  
           & $\mathbf{1.00}$ & $3.43 \pm 0.93$ & $4.01$ & $2.41$  
           & $\mathbf{1.00}$ & $\mathbf{2.54 \pm 0.23}$ & $3.26$ & $3.30$\\  

        D(L) & $\mathbf{1.00}$ & $0.73 \pm 0.61$ & $1.50$ & $1.48$
                & $0.92$ & $1.33 \pm 1.08$ & $3.25$ & $3.07$ 
                & $0.92$ & $3.37 \pm 0.78$ & $3.60$ & $2.03$  
                & $0.98$ & $2.60 \pm 0.40$ & $\mathbf{3.14}$ & $3.01$\\  

        D(LR) & $0.54$ & $2.76 \pm 2.17$ & $1.47$ & $1.47$
                    & $0.44$ & $3.61 \pm 1.58$ & $3.80$ & $3.20$ 
                    & $0.34$ & $4.73 \pm 0.42$ & $4.85$ & $2.34$ 
                    & $0.58$ & $3.54 \pm 0.84$ & $3.49$ & $3.10$\\ 

        D(RR) & $\mathbf{1.00}$ & $\mathbf{0.50 \pm 0.21}$ & $\mathbf{1.46}$ & $\mathbf{1.41}$ 
                    & $0.92$ & $1.42 \pm 1.09$ & $3.34$ & $\mathbf{2.97}$
                    & $\mathbf{1.00}$ & $\mathbf{3.16 \pm 0.60}$ & $\mathbf{3.17}$ & $\mathbf{1.84}$  
                    & $\mathbf{1.00}$ & $2.73 \pm 0.50$ & $3.21$ & $\mathbf{2.91}$ \\  
        \hline
    \end{tabular}
\end{minipage}

\end{figure*}

We perform a benchmark comparison in four environments: two from the robosuite~\cite{zhu2020robosuite} and two from the Manipulation with Viewpoint Selection (MVS) tasks \cite{sun2024comparative}. Fig. \ref{fig:simulation_environments} provides illustrations of the four environments.
The robosuite environments, Lift and 
Can, involve a Panda arm performing manipulation tasks, such as lifting a red cube and placing a can into a category. The state-action spaces for the robosuite tasks are the end-effector space with quaternions for rotation. The MVS tasks include opening a microwave, and the same pick-and-place scenario we used for previous ablation studies.  

The methods we evaluate in this benchmark comparison include Behavior Cloning (BC), HG-DAgger, and three variants of DRIFT-DAgger: one that uses LoRA, one that uses LoRA with rank scheduler, and one that uses rank modulation and rank scheduler. 
For methods utilizing rank scheduler, we apply the sigmoid decay function with steepness $\tau$ set to 0.5. We set $r_{\text{max}}$ and $r_{\text{min}}$ to 2048 and 256, respectively, based on the maximum rank of the diffusion policy and prior ablation study on the terminal rank. For all methods that use LoRA, we set the scaling factor $\alpha$ to 1.0. 
Experimental parameters related to the interactive mechanism, including the number of offline rollouts in $\mathcal{D}_B$, bootstrapping epochs $\mathcal{I}$, and online iterations $\mathcal{J}$, are provided in Table \ref{tab:exp_config}. 

During the online iterations, we save checkpoints every 20 iterations. Each checkpoint undergoes evaluation with 50 rollouts, and the success rate, mean and standard deviation of task duration, the number of expert labels, and cumulative training time are recorded as metrics for comparison. 
The success rate is used as the primary performance metric. The mean and standard deviation of task duration reflect how consistent and certain a trained policy is in completing a given task. A lower mean and standard deviation of task duration suggest that the policy is well-trained and converges better to the desired behavior. The number of expert labels, when considered alongside the other metrics, provides insight into the sample efficiency of a specific training method. For example, at the same level of success rate, a lower number of expert labels indicates better sample efficiency. The cumulative training time is for reflecting the training efficiency.

To fairly and efficiently evaluate different methods, for each scenario, we first train an expert policy using human-collected data from \citet{robomimic2021} and \citet{sun2024comparative} for robosuite and MVS tasks, respectively. The expert policy performs interventions when the cosine similarity between the learner actions and expert actions falls below a threshold, as detailed in Table \ref{tab:exp_config}. The thresholds are computed based on the mean cosine similarity between consecutive steps in the expert training datasets.

As shown in Fig. \ref{fig:simulation_environments} and Table \ref{tab:sim_scenarios}, DRIFT-DAgger variants demonstrate a pronounced reduction in cumulative training time compared to BC and HG-DAgger.  Additionally, all interactive IL methods exhibit superior expert sample efficiency compared to BC, as evidenced by higher success rates with respect to the number of expert labels. An exception is observed in the DRIFT-DAgger variant with LoRA and rank scheduler, where the merging and re-injection of LoRA adapters destabilize training due to the initialization of new trainable parameters. In contrast, the DRIFT-DAgger variant that instantiates LoRA adapters only once during the transition to the online phase, or the variant that combines rank modulation and rank scheduler, achieve performance and sample efficiency comparable to HG-DAgger, which consistently uses full ranks instead of reduced ranks. The benefits of interactive IL are more pronounced in tasks with longer durations. 
Furthermore, the policies learned with DRIFT-DAgger using the combination of rank modulation and rank scheduler, exhibit good stability and better convergence to the desired behavior of the task, as indicated by the relatively lower standard deviation in task duration.

\subsection{Batch Training Time}
\label{sec:abl_rm}
\begin{figure}[htbp]
\centering
\begin{minipage}{\columnwidth}
\centering
\includegraphics[width=\columnwidth]{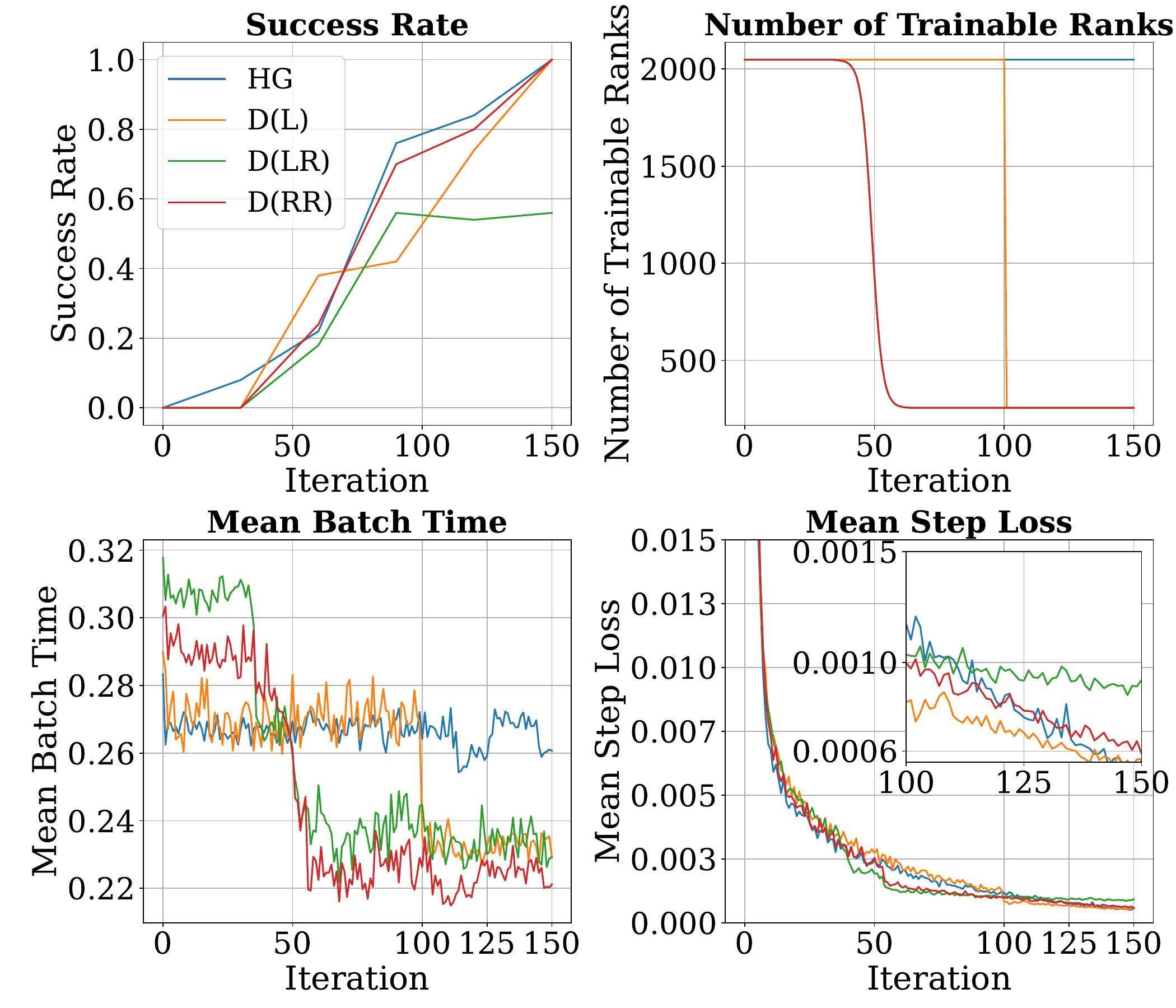}
\caption{Experimental results of three DRIFT-DAgger variants with HG-DAgger for demonstrating the reduced batch training time of DRIFT framework compared to full-rank training, while still maintaining equivalent performance.}
\label{fig:batch_training_time}
\end{minipage}

\vspace{1em} 

\begin{minipage}{\columnwidth}
\centering
\renewcommand\arraystretch{1.2}
\captionof{table}[Ablation on Training Strategies]{Summary of experimental results on mean batch training time (MBT) and success rate with full-rank and reduced-rank training methods.}
\label{tab:training_strategies}
    \begin{tabular}{l|c|c|c|c}
    \hline 
    Method & \makecell{Success \\ Rate} & \makecell{MBT \\ (Offline)} & \makecell{MBT \\ (Online)} & \makecell{MBT \\ (All Stages)} \\ 
    \hline
    HG             & $\mathbf{1.0}$ & 0.27 & 0.27 & 0.27 \\
    D(L)        & $\mathbf{1.0}$ & 0.27 & 0.23 & 0.26 \\
    D(LR)      & 0.56 & 0.27 & 0.23 & 0.26 \\
    D(RR)        & $\mathbf{1.0}$ & $\mathbf{0.26}$ & $\mathbf{0.22}$ & $\mathbf{0.24}$ \\
    \hline
    \end{tabular}

\end{minipage}
\end{figure}
To better understand and demonstrate the improved training efficiency of DRIFT-DAgger, we conduct an ablation study on the mean batch training time across all stages. The methods evaluated in this ablation study include HG-DAgger and three DRIFT-DAgger variants: one using LoRA, one combining LoRA with a rank scheduler, and one employing rank modulation alongside rank scheduler. We use HG-DAgger as the baseline for full-rank training.


This ablation study is performed using the MVS pick-and-place scenario, same as \S\ref{sec:abl_rs} and \S\ref{sec:abl_tr}. We use 100 offline demonstration rollouts, 100 offline bootstrapping epochs, and 50 online iterations for training. We set terminal ranks $r_{min}$ to 256 and use the sigmoid decay function with $\tau$ set to 0.5 for DRIFT-DAgger variants applied.

The results are presented in Fig. \ref{fig:batch_training_time} and Table \ref{tab:training_strategies}. We observe that the DRIFT-DAgger variant with fixed-rank LoRA and the variant with rank modulation and rank scheduler both achieve success rates comparable to the full-rank baseline, HG-DAgger. When considering batch training time, the variant with rank modulation and rank scheduler reduces the mean batch training time across all stages by 11\%, demonstrating improved training efficiency without sacrificing performance. Specifically, during the online training stage, this variant achieves an 18\% reduction in batch training time compared to the full-rank baseline. 

In contrast, the DRIFT-DAgger variant with LoRA and rank scheduler also shows reduced training time. However, the success rate significantly drops compared to HG-DAgger. This decline is attributed to the instability caused by merging and re-injecting LoRA adapters, which is also reflected in the higher step loss observed for this variant. Additionally, despite the use of a rank scheduler, the batch training time for this variant is slightly higher than that of the variant with rank modulation, likely due to the additional computational overhead introduced by LoRA within the DRIFT framework.

\begin{figure}
\centering
\includegraphics[width=\columnwidth]{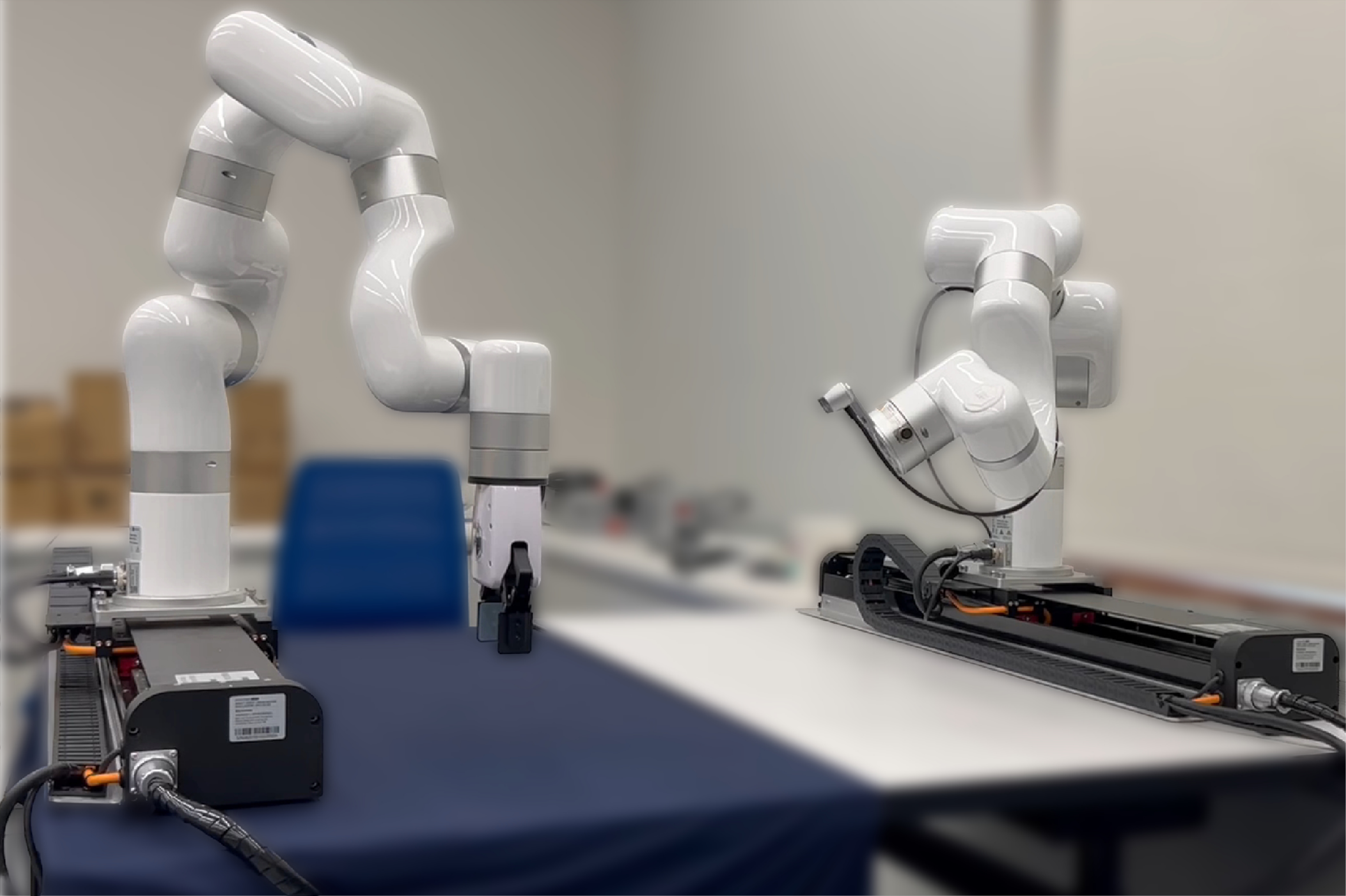}
\caption{The 17-DOF robotic system for real-world experiments aligns with the MVS simulation environments. It includes two xArm7 mounted on linear motors, with camera and gripper attached to each end effectors.}
\label{fig:real_world_system}
\end{figure}



%% file: 7-real_world_evaluation.tex
\section{Real-World Evaluation}
\label{sec:real_world_evaluation}

\label{sec:real_exp}

\begin{figure*}[htbp]
\centering
\begin{minipage}{\textwidth}
    \centering
    \includegraphics[width=\textwidth]{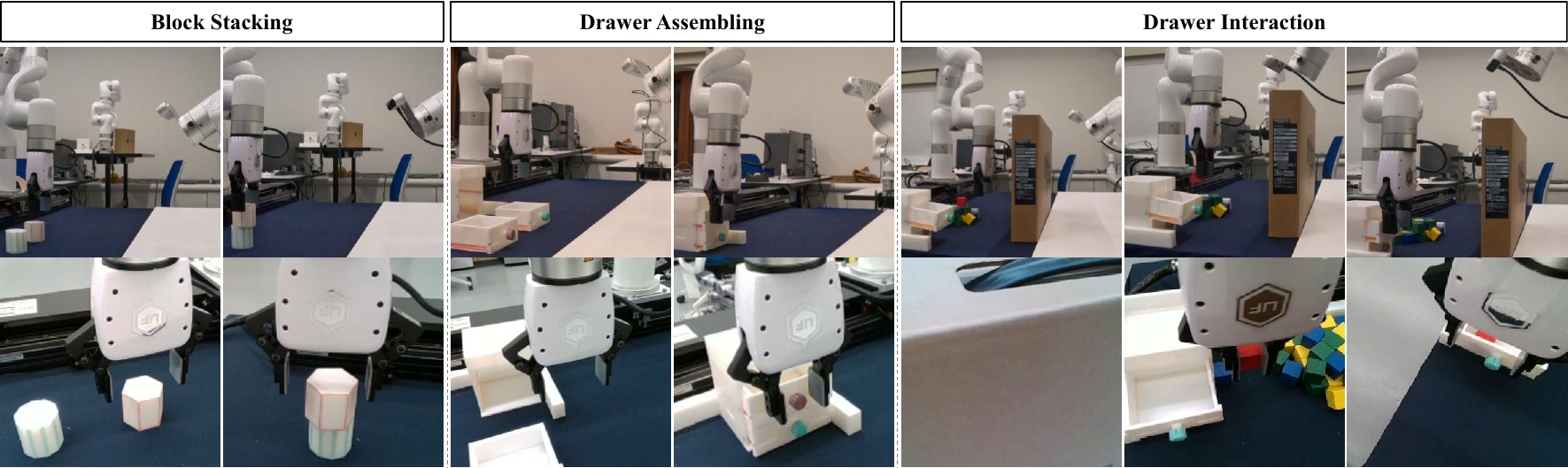} 
    \caption{The images show the tasks for real-world experiments. The upper row and lower row show the process of each task from a third-person perspective and a robot-perception perspective respectively.}
    \label{fig:real_world_scenarios}
\end{minipage}

\vspace{1em} 

\begin{minipage}{\textwidth}
\setlength{\tabcolsep}{5pt}
    \centering
    \renewcommand\arraystretch{1.05}
    \captionof{table}[Real-World Scenarios]{Summary of experimental results from real-world scenarios. The metrics include success rate (SR), mean and standard deviation of task duration (MSD), number of expert labels (NEL), and cumulative training time (CT). CT is measured in hours, MSD measured in minutes, and NEL is at the scale of $\times 10^{4}$.}
    \label{tab:real_scenarios}
    
    \begin{tabular}{l|cccc|cccc|cccc}
        \hline
        & \multicolumn{4}{c|}{Block Stacking} & \multicolumn{4}{c|}{Drawer Assembling} & \multicolumn{4}{c}{Drawer Interaction} \\
        \cline{2-13}
         & SR & MSD & NEL & CT & SR & MSD & NEL & CT & SR & MSD & NEL & CT \\
        \hline
        BC & $0.97$ & $\mathbf{0.41 \pm 0.04}$ & $4.86$ & $4.45$ 
            & $0.40$ & $1.76 \pm 0.21$ & $12.08$ & $14.50$
            & $0.83$ & $0.94 \pm 0.11$ & $4.82$ & $3.99$ \\ 

        HG & $\mathbf{1.00}$ & $0.42 \pm 0.04$ & $\mathbf{4.22}$ & $4.35$
            & $\mathbf{0.77}$ & $\mathbf{1.52  \pm 0.18}$ & $10.21$ & $14.21$
            & $0.90$ & $0.85 \pm 0.12$ & $\mathbf{3.94}$ & $3.87$\\ 

        D(L) & $\mathbf{1.00}$ & $0.43 \pm 0.05$ & $4.31$ & $4.19$
            & $0.67$ & $1.55  \pm 0.20$ & $10.29$ & $13.81$
            & $0.87$ & $0.89 \pm 0.11$ & $4.04$ & $3.73$\\ 

        D(LR) & $0.53$ & $0.58 \pm 0.05$ & $4.57$ & $4.23$
            & $0.20$ & $2.18 \pm 0.24$ & $12.11$ & $14.04$ 
            & $0.37$ & $1.09 \pm 0.18$ & $4.53$ & $3.79$\\ 

        D(RR) & $\mathbf{1.00}$ & $0.43 \pm 0.05$ & $4.25$ & $\mathbf{4.03}$ 
            & $0.73$ & $1.58 \pm 0.15$ & $\mathbf{10.01}$ & $\mathbf{13.28}$
            & $\mathbf{0.93}$ & $\mathbf{0.84 \pm 0.10}$ & $4.08$ & $\mathbf{3.59}$\\ 
        \hline
    \end{tabular}
    \vspace{-0.2cm}
\end{minipage}

\end{figure*}

To further validate the proposed DRIFT framework, we deploy DRIFT-DAgger in three real-world tasks, using a human teleoperator as the expert. The robotic system for these experiments, illustrated in Fig. \ref{fig:real_world_system}, mirrors the setup used in the simulated MVS tasks. The system comprises two UFactory xArm7 robots mounted on linear motors, with a gripper attached to one arm and a camera to the other. This configuration enables simultaneous manipulation and viewpoint adjustment. The state-action space is a reduced end-effector space, with the orientation of the camera automatically updated via an additional IK objective.

The real-world tasks, depicted in Fig. \ref{fig:real_world_scenarios}, include block stacking, which involves placing a non-cuboid block on top of another; drawer assembling, which involves inserting two drawer boxes into a drawer container; and drawer interaction, which requires the agent to first get rid of a visual occlusion (cardboard box), grasp a red cube from a cluttered area, place it in the drawer, and then close the drawer. The blocks and drawer components are 3D-printed using models from \cite{lee2021beyond} and \cite{heo2023furniturebench}. Training configurations for each task are detailed in Table \ref{tab:exp_config}, and the trained policies from each method are evaluated over 30 rollouts.

We use the success rate, mean and standard deviation of task duration, number of expert labels, and cumulative training time as metrics for the real-world experiments. The cumulative training time only measures the active computation during training, and not includes any other events between two training epochs, such as resetting the robot or recording the demonstration interactively. 

Results, summarized in Table \ref{tab:real_scenarios}, show trends consistent with the simulation experiments. All DRIFT-DAgger variants show noticeable reduction in cumulative training time. Interactive IL methods, except the DRIFT-DAgger variant with LoRA and rank scheduler, achieve similar or superior performance to BC, with improved sample efficiency, as indicated by a reduced number of expert labels. Despite using reduced-rank training, the other two DRIFT-DAgger variants perform comparably to HG-DAgger, which trains in a full-rank manner. The advantage of interactive methods becomes more pronounced for longer-duration tasks. For example, in the block stacking task, which has a mean duration of around 0.45 minutes, interactive methods improve sample efficiency by 11.32\% to 13.17\%. In contrast, for the drawer assembling task, with a mean duration of approximately 1.72 minutes, sample efficiency improves by 14.82\% to 17.14\%.

These real-world experiments validate that the DRIFT framework is effective in real-world settings, offering reduced training time, improved sample efficiency and robust performance despite employing reduced-rank training.

%% file: 8-discussion.tex
\section{Discussion}
\label{sec:discussion}

This work introduces DRIFT, a framework designed to leverage the intrinsic low-rank properties of large diffusion policy models for efficiency while preserving the benefits of overparameterization. To achieve this, we propose rank modulation and rank scheduler, which dynamically adjust trainable ranks using SVD and a decay function. We instantiate DRIFT within an interactive IL algorithm, DRIFT-DAgger, and show this efficacy of this method through extensive experiments and ablation studies in both simulation and real-world settings. Our results demonstrate that DRIFT-DAgger reduces training time and improves sample efficiency while maintaining performance on par with full-rank policies trained from scratch.

\subsection{Limitations}
This work evaluates and demonstrates the DRIFT framework using DRIFT-DAgger as an instantiation within the IL paradigm. However, prior to the adoption of large models, online reinforcement learning (RL) approaches \cite{schulman2017proximal, haarnoja2018soft} were also a popular area of research. This work does not explore the application of the DRIFT framework in the online RL paradigm. Investigating the potential of DRIFT within online RL could serve as an valuable direction for future research.

Additionally, while we have tested and evaluated various decay functions for the rank scheduler, the current implementation of dynamic rank adjustment in the DRIFT framework follows a monotonic schedule. Although we have conducted ablation studies on decay functions and terminal ranks, the impact of these design choices is likely task-dependent. 

Furthermore, the rank adjustment of different convolutional blocks in this work is applied uniformly throughout the U-Net backbone, even though different blocks may have varying highest possible ranks. Future research could explore more adaptive and intelligent strategies for adjusting trainable ranks to enhance training efficiency and performance, as well as identify suitable decay functions and terminal ranks for scenarios beyond those covered in this work.

\subsection{Implications}
As discussed in \S\ref{sec:related_works}, prior to the era of large models, innovations in robot learning primarily focused on learning processes with interaction. However, the increasing size of models has resulted in significantly longer training times, making many previous innovations in online interactive learning less practical due to the time required for policy updates. While the machine learning community has made progress in leveraging the intrinsic rank of large models to improve training efficiency, most of these methods are tailored for fine-tuning rather than training from scratch. This distinction arises from the availability of foundation models in general machine learning, whereas robotics often requires training policies from scratch to address scenario-specific tasks.

Despite years of research in the machine learning community, the concepts of overparameterization and intrinsic ranks remain relatively underexplored in robotics. This work introduces reduced-rank training as a means to address the challenges of training efficiency, thereby making online interactive learning methods more feasible in the era of large models for robot learning. By bridging these gaps, we aim to raise awareness within the robotics community about leveraging overparameterization and intrinsic ranks to design more efficient learning methods while preserving the powerful representations afforded by overparameterized models.


%% file: 9-appendix.tex
\clearpage

\section{Appendix}
\subsection{Additional Baselines and Variants}
\label{sec:appendix_additional_comp}

\begin{figure}[htbp]
\centering
\begin{minipage}{\columnwidth}
\centering
  \includegraphics[width=\linewidth]{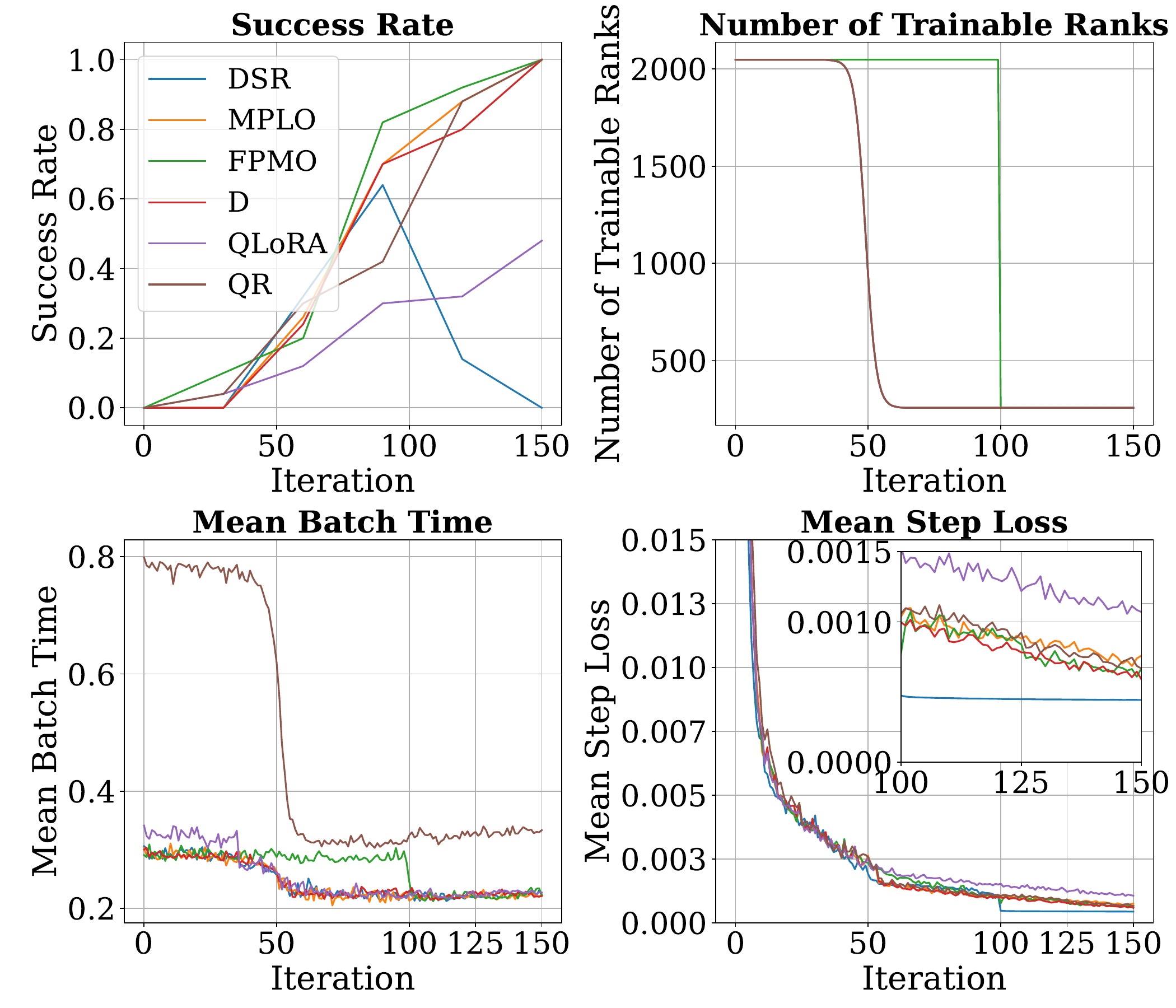}
  \caption{Results of additional baselines and variants on the MVS Pick and Place task, comparing DRIFT-DAgger (D) to QLoRA (sigmoid decay, steepness 0.5), and four variants: QR (QR decomposition per step), FPMO (full-rank offline, static RM with $r_{min}=256$ online), MPLO (RM offline, LoRA online), and DSR (fine-tuning last $10\%$ denoising steps online). }
  \label{fig:pnp}
\end{minipage}

\vspace{1em} 

\begin{minipage}{\columnwidth}
    \centering
    \renewcommand\arraystretch{1.2}
    \captionof{table}{Summary of experiments for additional baseline and variants on the PnP task. \label{tab:pnp}}
    \resizebox{\columnwidth}{!}{ 
    \begin{tabular}{l|c|c|c|c}
    \hline 
    \makecell{Method} & \makecell{Success \\ Rate} & \makecell{MBT \\ (Offline)} & \makecell{MBT \\ (Online)} & \makecell{MBT \\ (All Stages)} \\ 
    \hline
    D        & $\mathbf{1.0}$ & $\mathbf{0.26}$ & $\mathbf{0.22}$ & $\mathbf{0.24}$ \\
    QR    & $\mathbf{1.0}$ & 0.55 & 0.33 & 0.47 \\
    FPMO    & $\mathbf{1.0}$ & 0.28 & 0.22 & 0.26 \\
    MPLO    & $\mathbf{1.0}$ & 0.26 & 0.22 & 0.24 \\
    DSR    & 0.0 & 0.26 & 0.22 & 0.24 \\
    QLoRA    & 0.48 & 0.27 & 0.23 & 0.25 \\
    \hline
    \end{tabular}
    }
\end{minipage}
\end{figure}

In addition to the baselines considered in the paper, we also compare DRIFT-DAgger (D) with QLoRA~\cite{dettmers_qlora}, and four additional DRIFT variants: QR (using QR decomposition per gradient step,  FPMO (full-rank offline, static RM with $r_{min}$ = 256 online), MPLO (RM offline, LoRA online), and DSR (fine-tuning last 10\% denoising steps online).
The results on the MVS Pick and Place (PnP) task can be found in Table~\ref{tab:pnp} and Fig.~\ref{fig:pnp}.

First, we observe that QLoRA on the MVS PnP task shows inferior performance to ours.
Also, DIRFT-DAgger with QR decomposition reduces training efficiency significantly with negligible performance gains.
MPLO perform similarly to DRIFT-DAgger with RM, further suggesting DRIFT outperforms LoRA-based methods due to instability of LoRA from injection and reinstantiation, and its need for well pre-trained weights, whereas DRIFT suits robot learning trained from scratch.
FPMO demonstrates slightly worse training efficiency due to full-rank offline training.

DSR causes model collapse during online adaptation despite having lower loss due to overfitting the last few steps during denoising. This is likely because it also originates from diffusion models in image generation \cite{clark2023directly}, which often assume well pre-trained models that are unavailable in diffusion policy context for robot learning.

\subsection{Additional Experiments on the Choice of $r_{min}$ in the Microwave task}
\label{sec:appendix_microwave_rmin_sweep}

\begin{figure}[htbp]
\centering
\begin{minipage}{\columnwidth}
\centering
  \includegraphics[width=\linewidth]{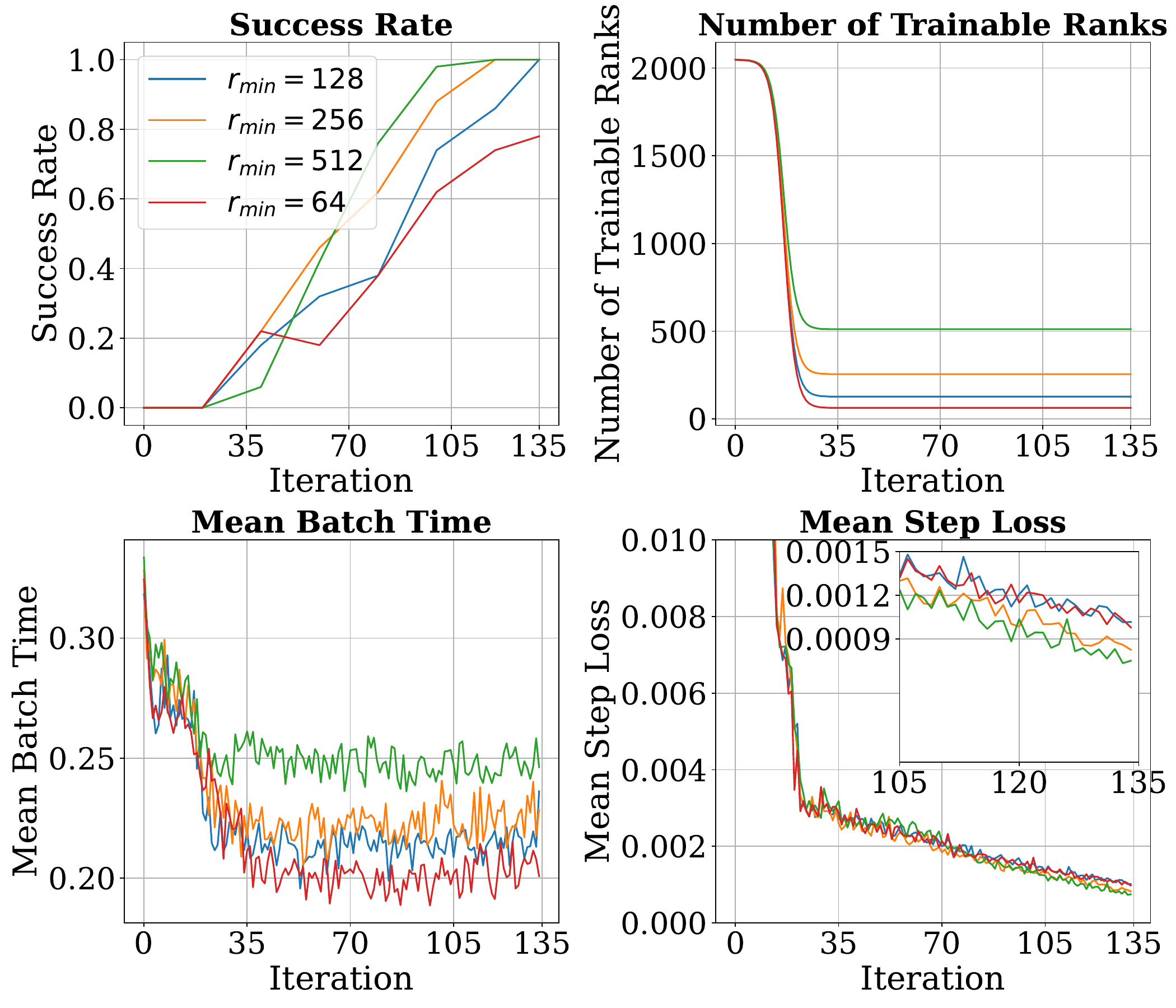}
  \caption{$r_{min}$ sweep on Microwave task. This task uses 100 offline rollouts, 35 offline epochs, and 100 online iterations to train, whereas the PnP task uses 100 offline rollouts, 100 offline epochs, and 50 online iterations for $r_{min}$ ablation.}
  \label{fig:microwave}
\end{minipage}

\vspace{1em} 

\begin{minipage}{\columnwidth}
    \centering
    \renewcommand\arraystretch{1.2}
    \captionof{table}{SSummary of $r_{min}$ sweep results on Microwave task.\label{tab:microwave}}
    \resizebox{\columnwidth}{!}{ 
    \begin{tabular}{l|c|c|c|c}
    \hline 
    \makecell{$r_{\text{min}}$} & \makecell{Success \\ Rate} & \makecell{MBT \\ (Offline)} & \makecell{MBT \\ (Online)} & \makecell{MBT \\ (All Stages)} \\ 
    \hline
    64     & 0.78 & $\mathbf{0.25}$ & $\mathbf{0.20}$ & $\mathbf{0.21}$ \\
    128    & $\mathbf{1.0}$ & $\mathbf{0.25}$ & 0.21 & 0.22 \\
    256    & $\mathbf{1.0}$ & 0.26 & 0.22 & 0.23 \\
    512    & $\mathbf{1.0}$ & 0.27 & 0.25 & 0.25 \\
    \hline
    \end{tabular}
    }
\end{minipage}
\end{figure}

We also conducted an $r_{min}$ sweep on the Microwave task, complementing the PnP task sweep in the paper. Fig. \ref{fig:microwave} and Table \ref{tab:microwave} show that, despite differing tasks and training configuration, the trend is similar to the PnP $r_{min}$ ablation: $r_{min}=256$ converges slightly faster than $r_{min}=128$. This suggests $r_{min}=256$ as a reliable heuristic for manipulation tasks with the specification of diffusion policies per \cite{chi_dp}.